\pdfoutput=1

\documentclass[11pt]{article}
\usepackage{acl}

\usepackage{times}
\usepackage{latexsym}
\usepackage[notext,notextcomp]{stix2}
\usepackage{tikz}
\usepackage{tikz-network}
\usepackage{arydshln}
\usepackage{nert}
\usepackage{colortbl}
\usepackage{lscape}

\usepackage[T1]{fontenc}

\usepackage[utf8]{inputenc}

\usepackage{microtype}

\definecolor{darkyellow}{rgb}{0.85,0.65,0}
\definecolor{darkgreen}{rgb}{0.05,0.6,0.05}

\usetikzlibrary{decorations,decorations.markings,decorations.pathmorphing,decorations.pathreplacing}
\usetikzlibrary{svg.path}

\pgfkeys{/tikz/.cd,
    contour distance/.store in=\ContourDistance,
    contour distance=-10pt, 
    contour step/.store in=\ContourStep,
    contour step=1pt,
}

\pgfdeclaredecoration{closed contour}{initial}
{%
\state{initial}[width=\ContourStep,next state=cont] {
    \pgfmoveto{\pgfpoint{\ContourStep}{\ContourDistance}}
    \pgfcoordinate{first}{\pgfpoint{\ContourStep}{\ContourDistance}}
    \pgfpathlineto{\pgfpoint{0.3\pgflinewidth}{\ContourDistance}}
    \pgfcoordinate{lastup}{\pgfpoint{1pt}{\ContourDistance}}
    
  }
  \state{cont}[width=\ContourStep]{
     \pgfmoveto{\pgfpointanchor{lastup}{center}}
     \pgfpathlineto{\pgfpoint{\ContourStep}{\ContourDistance}}
     \pgfcoordinate{lastup}{\pgfpoint{\ContourStep}{\ContourDistance}}
  }
  \state{final}[width=\ContourStep]
  { 
    \pgfmoveto{\pgfpointanchor{lastup}{center}}
    \pgfpathlineto{\pgfpointanchor{first}{center}}
  }
}

\title{Linguistic Frameworks Go Toe-to-Toe\\ at Neuro-Symbolic Language Modeling}

\author{Jakob Prange \quad Nathan Schneider \\
  Georgetown University \\
  {\textsmaller[0.5]{\{\emldisplay{jp1724@georgetown.edu}{jp1724}, \emldisplay{nathan.schneider@georgetown.edu}{nathan.schneider}\}\texttt{@georgetown.edu}}} \\\And
  Lingpeng Kong \\
  The University of Hong Kong \\
  {\textsmaller[0.5]{\emldisplay{lpk@cs.hku.hk}{lpk@cs.hku.hk}}}\\}

\begin{document}

\maketitle

\begin{abstract}
    We examine the extent to which, in principle, %
    different syntactic and semantic graph representations can complement and improve neural language modeling. 
    Specifically, by conditioning on a subgraph
    encapsulating the locally relevant sentence history, can a model make better next-word predictions than a pretrained sequential language model alone?
    With an ensemble setup consisting of GPT-2
    and ground-truth graphs from one of 7 different formalisms, we find that  
    the graph information indeed improves perplexity and other metrics.
    Moreover, this architecture provides a new way to compare different frameworks of linguistic representation. In our oracle graph setup, training and evaluating on English WSJ, \emph{semantic constituency} structures prove most useful to language modeling performance---outpacing syntactic constituency structures as well as syntactic and semantic dependency structures.
\end{abstract}

\section{Introduction}

Linguistic theories posit that humans can take advantage of hierarchical structure related to some notion of \emph{compositionality} to produce and comprehend utterances with complex meanings. 
Yet explicit representations of this kind of structure are harder to come by than raw text, and large-scale pretrained neural language models \citep[e.g.,][]{devlin-19-bert,radford2019language} have managed to perform strikingly well at contextually encoding and predicting words from distributional evidence alone.
At the same time, there are good reasons to doubt that these models can be said to \textit{understand} language in any meaningful way \citep{trott-etal-2020-construing,bender-koller-2020-climbing,merrill-etal-2021-provable}.
To address this conundrum, people have started to explore probing pretrained models \citep[][\textit{inter alia}]{liu2019linguistic,tenney-etal-2019-bert} and supplementing training data with linguistic structure guidance \citep[][\textit{inter alia}]{strubell-etal-2018-linguistically,swayamdipta-18,peng-etal-2019-palm,wu-2021-infusing}.

\begin{figure*}[ht]
    \centering\small
    \includegraphics[width=\textwidth]{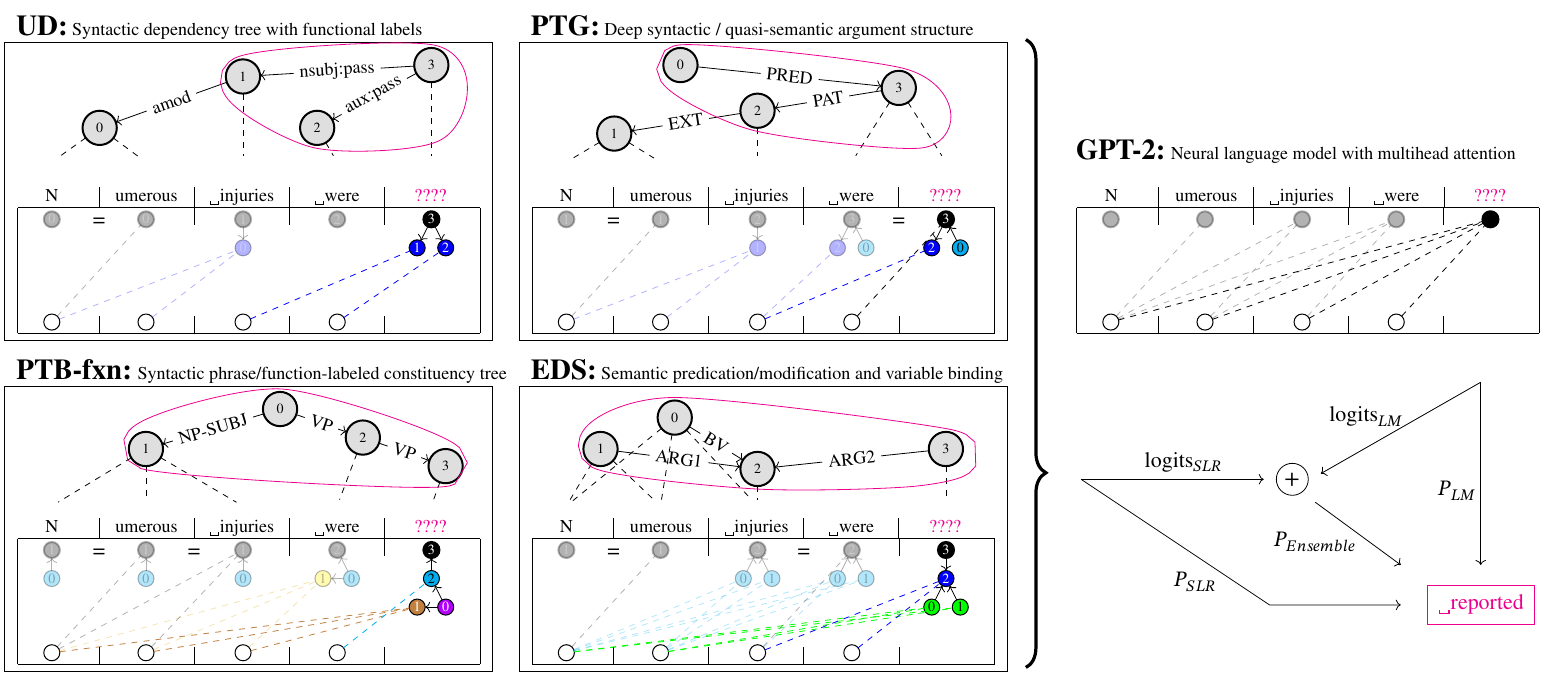}
    \caption{
    Contrasting GPT-2's incremental attention mechanism (top right) with incremental context \textit{slices} obtained from linguistic graphs (left four panels) of four different formalisms (\cref{sec:formalisms}).
    As shared tokenization we use GPT-2's byte-pair encoding. 
    Slice nodes are color-coded by local relation type (\textcolor{black}{black: target}, \textcolor{cyan}{cyan: parent}, \textcolor{blue}{blue: child}, \textcolor{darkgreen}{green: coparent}, \textcolor{darkyellow}{yellow: sibling}, \textcolor{purple}{purple: grandparent}, \textcolor{brown}{brown: aunt}). Dashed lines indicate the token anchoring of original (big grey) graph nodes and, correspondingly, which previous tokens (empty circles) are accessible for each next-token prediction.
    In the bottom right we visualize how different models arrive at their prediction (\cref{sec:motivation,sec:p-slr}).
    }
    \label{fig:slice}\label{fig:slr}
\end{figure*}

A question that has received less attention is \emph{which kind} of symbolic linguistic representation (SLR) is most conducive to guiding neural language models (LMs).
Numerous domain-general candidates exist \citep{abend-17,oepen-etal-2019-mrp,oepen-etal-2020-mrp,zabokrtsky-20,muller2020grammatical}: some are focused
on syntactic structure, others on semantics (\cref{sec:slr}; big grey example graphs in the left panels of \cref{fig:slr}).
Frameworks vary along several dimensions, with different label inventories and treatments of specific constructions. 
Formal differences include the type of structure (dependency or constituency, one or multiple parents, projectivity)
and its relation to the input string. 
In general, different design choices may aim to capture different kinds of generalizations or facilitate different kinds of processing, and may make parsing raw text easier or harder. 
It is often not obvious which framework should be chosen for best results on an external task---or indeed, how to even perform a controlled comparison across frameworks.

In this paper we investigate whether structurally guided language modeling can serve as a benchmark task for directly comparing linguistic representations. 
Specifically, we evaluate on next-word prediction---a relatively neutral task in that it does not
rely on any artificial test suite, nor does it target a specific downstream application where one linguistic framework may have an advantage.\footnote{Our findings are limited to a particular \emph{language} (English) and \emph{domain} (financial news) in which gold graphs from multiple frameworks are available for the same sentences, but such annotations could be obtained for other samples in the future.}

We devise a method for selecting and encoding partial views of linguistic graphs over the preceding context relevant to predicting the next token (\cref{sec:motivation,sec:method}).\footnote{Our code is available to the research community at \url{https://github.com/jakpra/LinguisticStructureLM}.} We call these views \textit{slices} (small per-token graphs and dashed lines in \cref{fig:slice}).
Our neuro-symbolic encoder statically allocates distinct vector dimensions for different structural relations within each slice, which in the incremental setting is much faster and more flexible than computation-intensive deep graph encoders.
Using this encoding, we compare 7~SLR formalisms by virtue of their incremental language modeling capability in an controlled experimental setup (\cref{sec:lm-exp}) on
jointly annotated ground-truth data
\citep{oepen-etal-2019-mrp,oepen-etal-2020-mrp}.
The results (\cref{sec:lm-results}) suggest that linguistic graphs are indeed informative for next-word prediction, complementing what is learned in pretraining. This invites future research quantifying different formalisms' design choices (\cref{sec:discussion}).

\section{Background: Symbolic Linguistic Representation}\label{sec:slr}

Following a long tradition in formal linguistics, graph-structured representations of language qualitatively describe grammatical and logical relations among words. %
The SLR paradigm has recently seen a revival in the form of larger-scale \textit{treebanking} and \textit{sembanking} for training neural parsers.

Formally, an SLR instance is a directed acyclic graph (DAG) $G = \langle V, E, \alpha\rangle$, with vertices $V$, labeled edges $E$, and an anchoring function $\alpha : V \rightarrow \mathbf{w}$ that maps each vertex to a (potentially empty) subset of tokens in the sentence.
We broadly distinguish SLR frameworks along two dimensions:\footnote{
See \citet{abend-17,koller-etal-2019-graph,prange-etal-19-semantically} for more detailed taxonomies.}
\paragraph{Scope.} A main goal of \textit{syntactic} representations is to explain distributional patterns in word order; they tend to be rooted trees with often projective anchoring functions.
\textit{Semantic} formalisms are meaning-oriented, aiming to capture the higher-level logic expressed in a sentence; thus, they may have more complex structures, including reentrant edges and discontiguous anchors.
\paragraph{Structure.} SLRs can further be subdivided into \textit{dependency} and \textit{constituency} structures.
The former are relatively shallow, while the latter contain abstract nodes with no or multiple word anchors.

\section{Overview: Language Modeling with Linguistic Graphs}\label{sec:motivation} 

Our main goal is to quantify the predictive power of different SLRs by combining them with a pretrained language model and measuring how this affects next-token generation performance.
A language model (LM) assigns probabilities to sentences and can be used to both process existing sentences and generate new ones.
As is standard practice, we treat sentences as length-$n$ sequences of word tokens, $\mathbf{w} = \langle {w}_0, {w}_1, \dots, {w}_{n-1}\rangle$.
An \textit{incremental} LM factorizes the joint probability of the sentence in terms of the probability of each word $w_i$ conditioned on previous tokens $\mathbf{w}_{<i}$; \cref{eq:lm}.

\finalversion{
This can be approximated with a Markov assumption, where the history is limited by window size $m$: 
\begin{align}
\begin{split}
P_{LM}(\mathbf{w}) 
= &\ \prod_{i=0}^{n-1} P_{LM}(w_i|\mathbf{w}_{<i}) \\
\approx &\ \prod_{i=0}^{n-1} P_{LM}(w_i|w_{i-m}, \dots, w_{i-1})
\end{split}
\end{align}
}

Here we describe at a high level how we process (oracle) SLR graphs for use in this language modeling scenario, i.e., to obtain context-conditional vocabulary distributions from them.
In contrast to sequential LMs, contexts are now graph-structured, and which context tokens to select as well as in what way they are related to the target token is determined by the underlying SLR graph $G$; \cref{eq:slr}.
\begin{align}
P_{LM}(\mathbf{w}) & = \prod_{i=0}^{n-1} P_{LM}(w_i|\mathbf{w}_{<i}) \label{eq:lm} \\
P_{SLR}(\mathbf{w}) & := P(\mathbf{w}|G) \label{eq:slr}
\end{align}

This general idea is closely related to syntactic language modeling \citep[][\textit{inter alia}]{pauls-klein-2012-large,gubbins-vlachos-2013-dependency}. 
We extend this line of work to arbitrarily complex syntactic and semantic DAG structures and, in doing so, take particular care to restrict conditioning contexts from accessing not only future \textit{words} but also future \textit{subgraphs}, so effectively top-down and left-to-right.
Our procedure is as follows:

First, we select for each token position $i$ to be predicted a subgraph $G_i$, called the token's \textbf{slice}. Slices are both \textit{admissible} in the language modeling setting, i.e., they do not violate the left-to-right conditioning order, and \textit{relevant} to the token prediction according to some criteria---here we consider criteria based on structural relationships generally, without relying on formalism-specific labels (\cref{sec:slice}).
Consider the small colored subgraphs for each token in \cref{fig:slice}: the EDS-slice for the target `reported', for example, starts at node 3, and extends to the ARG2-child 2, ARG1-coparent 1, and BV-coparent 0, which are anchored, respectively, in the spans `injuries', `Numerous', and `Numerous injuries'). Recall from \cref{sec:slr} that context words $\mathbf{w}_{<i}$ are contained in $G_i$, to the extent that they are anchored in a node reachable from $w_i$.
Inspired by Markov assumptions of independence in generative modeling and 
Markov blankets in causal \finalversion{Bayesian }networks, SLR graph slicing thus allows us to factorize $P(\mathbf{w}|G)$ as
\begin{align}
\begin{split}
P(\mathbf{w}|G)   
:=  \prod_{i=0}^{n-1} P(w_i|G_i).
\end{split}
\end{align}

Next, we \textbf{encode} each graph slice as a fixed-sized vector. Prior approaches to encoding linguistic graphs for neural modeling have involved serialization, e.g., as parser transition sequences \citep[][\textit{inter alia}]{qian-etal-2021-structural}, recursive auto-encoders \citep{tai-15,roth-lapata-2016-neural}, and graph-convolutional networks \citep[GCNs;][]{yang_deng_2020_incremental,wu-2021-infusing}.
However, transition sequences for non-tree graphs are subject to spurious ambiguity; and we find that graph-structured neural networks are impractical in the incremental setting (\cref{sec:gnn-baseline}).
Instead, we propose a computationally inexpensive method for statically and deterministically projecting slices into a high-dimensional space by vector concatenation (\cref{sec:snap}).

Finally, we compute output \textbf{distributions} $P(W_i|G_i)$ from the vector representations (\cref{sec:p-slr}).

\section{Modeling Details}\label{sec:method}

\subsection{Slicing Graphs}\label{sec:slice}

A slice $G_{i}$ is a connected subgraph of $G$ that captures $w_i$'s linguistically structured context, masking $w_i$ itself (or else estimating $P(w_i|G_{i})$ would be trivial).
$G_{i}$ always minimally consists of $w_i$'s \textit{direct anchor} node $a_i = \operatorname{Select}(\{v : w_i \in \alpha(v)\})$.
Starting from $a_{i}$, we traverse the graph and add vertices and edges that are connected to $a_{i}$ via paths of a few specific relative types, REL. Here we settle on 6 types: parents, siblings, grandparents, parents' siblings, children, and coparents.
The vertices $V_i$ and edges $E_i$ for slice $G_i = \langle V_i, E_i, \alpha \rangle$ consist then of the union of these sets.\footnote{See \cref{app:anchor,app:rels} for details.}

To prevent information leakage from future tokens, we discard from $G_i$ all nodes $\{v: \alpha(v) = w_j, j > i\}$ which are \textit{only} anchored in tokens \textit{following} $w_i$. E.g., in \cref{fig:slice}, the UD-slice for the token `were' does not contain the parent node 3 because that is anchored only in the following token `reported' (and thus the sibling 1 cannot be accessed either). If a node's anchors \textit{contain or overlap with} $w_i$ (i.e., the node is a non-terminal above $w_i$), we retain the node and its edges but remove its token anchors.

\finalversion{Note that two distinct slices $G_i, G_j, i \neq j$ may overlap in the original graph but can always be distinguished by their structural identity with respect to the anchor.}

\subsection{Vectorizing Graph Slices}\label{sec:snap}

Because slices can be large, we partition each slice's nodes by structural \textit{relative type}, in order to aggregate them into a fixed-length summary vector.
Specifically, we allocate capacities for each relative type: $\gamma_{\text{rel}}=2$ for parents, siblings, aunts, and children, and 1 for grandparents and coparents.
Up to $\gamma - 1$, relative nodes $V_{\text{rel}}$ are added `with high resolution', maintaining their identity and order; beyond the capacity, relatives are aggregated `with low resolution'; \cref{eq:resol}. 
Within each relative type, precedence $k$ is given to relatives whose token anchors are sequentially closer to $w_i$.

\begin{align}\label{eq:resol}
\begin{split}
    \text{HiRes}_{i, \text{rel}} = &\ \left\langle v_{\text{rel},k} : k < \gamma_\text{rel} \right\rangle \\
    \text{LoRes}_{i, \text{rel}} = &\ \{v_{\text{rel},k} : k \geq \gamma_\text{rel}\}
\end{split}
\end{align}
\finalversion{\begin{align}\label{eq:hyper}
\begin{split}
    \hat{V}_{i} = &\ \bigcup_{\text{rel}}\{\{v\} : v \in \text{HiRes}_{i, \text{rel}}\} \cup \left\{\text{LoRes}_{i, \text{rel}}\right\} \\
    \hat{G}_{i} = &\ \left\langle \hat{V}_i, E_i, \alpha \right\rangle
\end{split}
\end{align}
}

Next we look up the relatives' edge label and word vector encodings\footnote{See \cref{app:enc} for details.} $\vec{l}_k$ and $\vec{w}_k$ and collate them into a single vector $\overrightarrow{s}_{i,\text{rel}}$ per relative type.
High-resolution vectors are concatenated $\bigoplus$ and low-resolution vectors are averaged; \cref{eq:concat-res}.
Finally, we concatenate all of these (zero-padded) relative-vectors to obtain the final vector representation of the whole slice, $\overrightarrow{s}_i$; \cref{eq:concat-all}.
At a high level, this vector essentially specifies a deterministic, structured, typed, discrete self-attention over the token history.

\begin{align}\label{eq:concat-res}
\begin{split}
    \overrightarrow{s}_{i,\text{rel}}^{\text{HiRes}} & =\ \bigoplus_{k \in \text{HiRes}_{i, \text{rel}}} \left[\vec{l}_k ; \vec{w}_k\right] \\
    \overrightarrow{s}_{i,\text{rel}}^{\text{LoRes}} & =\ \sum_{k \in \text{LoRes}_{i, \text{rel}}} \frac{\left[\vec{l}_k ; \vec{w}_k\right]}{|\text{LoRes}_{i, \text{rel}}|_+} 
\end{split}
\end{align}
\begin{align}\label{eq:concat-all}
\begin{split}
    \overrightarrow{s}_i = &\ \bigoplus_{\text{rel} \in \text{REL}} \left[\overrightarrow{s}_{i,\text{rel}}^{\text{HiRes}} ; \overrightarrow{s}_{i,\text{rel}}^{\text{LoRes}}\right]
\end{split}
\end{align}

\subsection{Predicting Emission Distributions}\label{sec:p-slr}

We compute model posteriors for next-token predictions as
\begin{align*}
P_\mu(w_i=v_k|\operatorname{context}_{i,\mu}) = &\ \operatorname{SoftMax}(\operatorname{logits}_{i,\mu})[k],
\end{align*}
where $\mu$ is either a pure SLR model or LM, or an ensemble of the two (bottom right of \cref{fig:slice}).

\paragraph{SLR only.}
As described above, we define $\operatorname{context}_{i,SLR}$ as $G_i$, which is encoded as $\overrightarrow{s_i}$.
We obtain $P_{SLR}$ by letting the slice-vectors serve as inputs to a $d$-multilayer perceptron (MLP) with a final softmax layer over the vocabulary, which yields the estimated token emission distributions.
\begin{align*}
\operatorname{logits}_{i,SLR} = &\ \operatorname{MLP}_d(\overrightarrow{s_i}) \\
\operatorname{MLP}_d(x) = &\ H^{(d)}\left( \dots H^{(1)}(x) \right) \text{Emb}^\top,
\end{align*}
where \text{Emb} is an embedding matrix.

\paragraph{LM + SLR.}
Since we want to measure whether and how much the information contained in the SLR can contribute to state-of-the-art language models, our primary experimental condition is a combined setup $P_{\textit{Ensemble}}$, where logits obtained from slice-encodings are added to a base neural LM's logits before taking the softmax:
\begin{align*}
\operatorname{logits}_{i,\textit{Ensemble}} = &\ \operatorname{logits}_{i,SLR} + \operatorname{logits}_{i,{LM}}, \\
\text{with\hspace*{1cm}}\operatorname{logits}_{i,{LM}} = &\ \operatorname{LM}(\mathbf{w}_{< i}).
\end{align*}

\paragraph{LM only.}
$P_{LM}$, i.e., the bare LM without any exposure to SLR graphs, serves as a baseline.

\section{Experimental Setup}\label{sec:lm-exp}

All models are implemented in PyTorch and experiments are run on 1 NVIDIA Tesla T4 GPU. Model hyperparameters are reported in \cref{app:hyper}.

\subsection{Data}\label{sec:lm-data}

\begin{table}[t]
    \centering\small
    \setlength{\tabcolsep}{4.8pt}
    \begin{tabular}{lrrr}
        & \multicolumn{1}{c}{\textbf{Sentences}} & \multicolumn{1}{c}{\textbf{Tokens}} & \multicolumn{1}{c}{\textbf{Vocabulary}} \\\toprule
        Train & 26,325 & 658,475 & 27,344  \\
        \hspace*{1em}Train (EarlyStop) & 23,692 & 591,829 & 26,422  \\
        \hspace*{1em}Dev (EarlyStop) & 2,633 & 66,646 & 10,073 \\
        Eval & 921 & 22,596 & 5,364 \\\bottomrule
    \end{tabular}
    \caption{Data statistics.}
    \label{tab:data}
\end{table}

Our dataset consists of the intersection of Wall Street Journal (WSJ; English financial news) sentences that have been annotated with syntactic trees in the Penn Treebank \citep[PTB;][]{marcus-etal-1993-ptb,hovy2006ontonotes}\footnote{\scriptsize\url{https://catalog.ldc.upenn.edu/LDC2013T19}} as well as a range of semantic representation formalisms for the MRP 2019 \& 2020 shared tasks \citep{oepen-etal-2019-mrp,oepen-etal-2020-mrp}. 
Summary statistics are shown in \cref{tab:data}. Our preprocessing steps are described in \cref{app:preproc}.

\subsection{SLR Formalisms}\label{sec:formalisms}

The 7 (versions of) linguistic representation frameworks examined in this study are listed in \cref{tab:main_results}, along with their classifications along the scope and structure dimensions.
We draw the structural dependencies vs.~constituencies distinction (described at a high level in \cref{sec:slr}) based on specific properties of the MRP shared task data: a framework is considered a dependency framework if all edges are only between pairs of individual word anchors at a time; if there are any unanchored\footnote{Not including ``ROOT'' nodes in UD.} nodes or nodes anchored in more than one linguistic word token, it is considered a constituency framework.\footnote{See \cref{app:const-dep} for details.} Below we give a brief description of each framework.

\textbf{PTB} trees specify hierarchically nested syntactic constituents.
We consider two labeling variants: basic phrase structure (\textbf{-phr}) and phrase types refined with functional specifications (\textbf{-fxn}).

Universal Dependencies \citep[\textbf{UD};][]{nivre-16,nivre-etal-2020-universal,de_marneffe-21} is a syntactic dependency representation with coarse, cross-linguistically applicable
edge labels.

DELPH-IN MRS Bi-Lexical Dependencies \citep[\textbf{DM};][]{ivanova-etal-2012-contrastive} and Elementary Dependency Structures \citep[\textbf{EDS};][]{oepen-lonning-2006-discriminant} are derived from underspecified logical forms computed by the English Resource Grammar \citep[][]{flickinger2000building,copestake2005minimal}.

Prague Semantic Dependencies \citep[\textbf{PSD};][]{hajic-etal-2012-announcing} and Prague Tectogrammatical Graphs (\textbf{PTG}) are syntactico-semantic predicate–argument structures converted from the Prague Functional Generative Description \citep[][]{sgall1986meaning,bohmova-03,hajic-etal-2012-announcing}.

\subsection{Language Model}\label{sec:lm}
The base language model we use in all our experiments is GPT-2 \citep[][as distributed in the huggingface-transformers PyTorch library]{radford2019language}. GPT-2 is a Transformer model \citep{vaswani-17} pretrained on a diverse collection of web texts\finalversion{, notably excluding Wikipedia articles}.
In contrast to other widely-used Transformers like BERT \citep{devlin-19-bert}, which optimize bidirectional masked language modeling, GPT-2 is incremental, i.e., next-word decisions only take into account the \textit{preceding} context.

\subsection{Training}

We train all models for 10 epochs with the AdamW optimizer \citep{loshchilov-19}, minimizing cross-entropy between the model posterior and the ground truth at each token position.

We perform early stopping with the last 10\% of the original training corpus set aside for development scoring after each epoch.\footnote{The R-GCN baseline (\cref{tab:rgcn_results}) is always trained for the full 10 epochs, but to ensure fairness, it is also only compared to concatenation-based encoders that have been trained for the full 10 epochs, too.} We keep the model state that achieves the best perplexity on the dev set. Peak development performance is reached after $\approx$3 epochs for SLR models, whereas finetuning GPT-2 by itself takes between 7 and 9 epochs.

\subsection{Evaluation}

We compute model perplexity (PPL) as the most standard language modeling evaluation measure, as well as accuracy (Acc) and confidence (Conf) of a model's top-ranked guess, mean reciprocal rank of the correct answer (MRR), and entropy of the model's token prediction posterior (H\finalversion{($W_i^\mu$); in contrast, PPL is the exponential of the \textit{cross}-entropy between model posterior and the ground-truth answer, H($W_i^\mu, w_i^*$), which is also the training loss}).
All metrics are reported as microaverages over the evaluation data at the BPE token level.\footnote{We compute average PPL over all sentences $j$ by exponentiating last: $\operatorname{exp}\left(\frac{1}{\sum_j|\textbf{w}_j|}\sum_j\sum_{i=0}^{|\textbf{w}_j|-1}-\operatorname{log}P_\mu\left(w_{ji}=v_{ji}^*|\text{context}_{ji,\mu}\right)\right)$}

\begin{table*}[t]
    \centering\small
    \setlength{\tabcolsep}{4.5pt}
    \begin{tabular}{l ccc ccH  c c c c c}
    & & & & \multicolumn{2}{c}{Training Efficiency} & & \multicolumn{5}{c}{Language Model Quality}\\
    \cmidrule{5-6}\cmidrule{8-12}
        \multicolumn{1}{l}{\textbf{Model}} & \multicolumn{2}{c}{\textbf{Scope/Struct}} & \multicolumn{1}{c}{\textbf{\#Labels}} & \multicolumn{1}{c}{\textbf{Speed} \textuparrow} & \multicolumn{1}{c}{\textbf{Size} \textdownarrow} & \multicolumn{1}{H}{\textbf{|Slice|}} & \multicolumn{1}{c}{\textbf{PPL} \textdownarrow} & \multicolumn{1}{c}{\textbf{H} [nats] \textdownarrow} & \multicolumn{1}{c}{\textbf{Acc} [\%] \textuparrow} & \multicolumn{1}{c}{\textbf{Conf} [\%] \textuparrow} & \multicolumn{1}{c}{\textbf{MRR} \textuparrow}  \\\toprule
        GPT-2 & & & & -- & \multirow{2}{*}{124.4M} & -- & 59.3	\phantom{\scriptsize $\pm .00$} & 4.09 \phantom{\scriptsize $\pm .000$} & 30.0 \phantom{\scriptsize $\pm .00$} & 31.2 \phantom{\scriptsize $\pm .00$} & .403	\phantom{\scriptsize $\pm 0.0$e-3} \\
        + Domain & & & & 15 & & -- & 45.9	\textcolor{gray}{\scriptsize $\pm .09$} &	3.64	\textcolor{gray}{\scriptsize $\pm .008$} &	33.3	\textcolor{gray}{\scriptsize $\pm .02$} &	34.9	\textcolor{gray}{\scriptsize $\pm .07$} &	.435	 \textcolor{gray}{\scriptsize $\pm \phantom{0}.3$e-3} \\\midrule
             &     &     &                &   &        &       & ***                             &                &            ***  \\
        + UD & syn & dep & \phantom{0}39 & 14 & $+$54.1M & 13.7K & 32.7 \textcolor{gray}{\scriptsize $\pm .18$} & 3.30 \textcolor{gray}{\scriptsize $\pm .013$} & 39.1 \textcolor{gray}{\scriptsize $\pm .15$} & \textbf{40.1} \textcolor{gray}{\scriptsize $\pm .14$} & .486 \textcolor{gray}{\scriptsize $\pm 1.2$e-3} \\
             &     &     &                &   &        &       & ***                             &                &           $-$ \\
        + DM & sem & dep & \phantom{0}59 & 15 & $+$54.4M & 14.0K & 31.4 \textcolor{gray}{\scriptsize $\pm .08$} & 3.24 \textcolor{gray}{\scriptsize $\pm .026$} & 38.9 \textcolor{gray}{\scriptsize $\pm .10$} & 40.2 \textcolor{gray}{\scriptsize $\pm .37$} & .491 \textcolor{gray}{\scriptsize $\pm \phantom{0}.6$e-3}  \\
             &     &     &                &   &        &       & ***                             &                &           $-$  \\
        + PSD & sem & dep & \phantom{0}90 & 16 & $+$54.9M & 14.5K & 30.7 \textcolor{gray}{\scriptsize $\pm .09$} & 3.21 \textcolor{gray}{\scriptsize $\pm .014$} & 39.1 \textcolor{gray}{\scriptsize $\pm .11$} & 40.9 \textcolor{gray}{\scriptsize $\pm .11$} & .491 \textcolor{gray}{\scriptsize $\pm \phantom{0}.5$e-3} \\
             &     &     &                &   &        &       & **                              &                &            ***  \\
        + PTB-phr & syn & const & \phantom{0}38 & 14 & $+$54.1M & 13.7K & 29.8 \textcolor{gray}{\scriptsize $\pm .18$} & 3.14 \textcolor{gray}{\scriptsize $\pm .029$} & 41.2 \textcolor{gray}{\scriptsize $\pm .19$} & 42.8 \textcolor{gray}{\scriptsize $\pm .42$} & .507 \textcolor{gray}{\scriptsize $\pm 1.3$e-3}  \\
             &     &     &                &   &        &       & ***                             &                &           *   \\
        + PTB-fxn & syn & const & 537 & 14 & $+$62.7M & 22.1K & 29.0 \textcolor{gray}{\scriptsize $\pm .28$} & 3.07 \textcolor{gray}{\scriptsize $\pm .049$} & 42.0 \textcolor{gray}{\scriptsize $\pm .30$} & 43.8 \textcolor{gray}{\scriptsize $\pm .60$} & .514 \textcolor{gray}{\scriptsize $\pm 1.8$e-3}  \\
             &     &     &                &   &        &       & ***                             &                &           *   \\
        + PTG & sem & const & \phantom{0}72 & 15 & $+$54.6M & 14.3K & 26.8 \textcolor{gray}{\scriptsize $\pm .26$} & 3.03 \textcolor{gray}{\scriptsize $\pm .041$} & \textbf{43.1} \textcolor{gray}{\scriptsize $\pm .12$} & 44.6 \textcolor{gray}{\scriptsize $\pm .51$} & .522 \textcolor{gray}{\scriptsize $\pm \phantom{0}.9$e-3}  \\
             &     &     &                &   &        &       & ***                             &                &           $-$  \\
        + EDS & sem & const & \phantom{0}10 & 15 & $+$53.6M & 13.3K & \textbf{24.7} \textcolor{gray}{\scriptsize $\pm .28$} & \textbf{2.92} \textcolor{gray}{\scriptsize $\pm .048$} & \textbf{43.1} \textcolor{gray}{\scriptsize $\pm .17$}  & 45.0 \textcolor{gray}{\scriptsize $\pm .55$} & \textbf{.527} \textcolor{gray}{\scriptsize $\pm 1.3$e-3}  \\\bottomrule
    \end{tabular}
    \caption{
    Main results: performance of language models combined with 7 SLR formalisms of different scope, structure, and label set (each corresponding to a $P_{\textit{Ensemble}}$ in \cref{sec:p-slr}), compared to vanilla GPT-2 and a version of GPT-2 that has been domain-finetuned on the raw text of the SLR training corpus ($P_{LM}$). We report each quality metric as mean $\pm$ stdev over 3 random seeds. We also report model size in \#parameters (all non-baseline models as absolute difference to baseline) and training speed in sentences per second as measures of efficiency. Statistical significance of the PPL and Acc differences to the next-best model (always adjacent rows) is reported as ***$p<.0001$ / **$p<.001$ / *$p<.005$ / $-$not significant (approximate randomization test as described in \citet{riezler-maxwell-2005-pitfalls}, with $R=$10,000 shuffles). We only consider a difference significant if $p<\alpha$ for all three random model initialization seeds. Best results in each column are \textbf{bolded}. For confidence, `best' means best-calibrated, i.e., the smallest relative difference to accuracy.
    }
    \label{tab:main_results}
\end{table*}

\section{Findings}\label{sec:lm-results}

\subsection{Main Results}

The most striking observation in terms of overall model performance (\cref{tab:main_results}) is that ground-truth linguistic graphs of all investigated linguistic formalisms \textit{improve} vanilla GPT-2 by a large margin, in all metrics. This improvement holds up when compared to a version of GPT-2 that is exposed to the raw WSJ text without the graphs; with this condition we control for mere domain differences between our evaluation data and the data GPT-2 was trained on originally (`$+$Domain' in \cref{tab:main_results}).
The large performance gap suggests that at least a subset of the oracle knowledge about linguistic structure is \textbf{not yet encoded} in the base language model, which learns from only raw text.

We observed that if we keep training for the entirety of 10 epochs, rather than early stopping based on development performance, we somewhat overfit to the training set. While accuracy itself is not affected very much by this, the models become increasingly overconfident ($\frac{\text{overall confidence}}{\text{overall accuracy}}$, which gets up to 8--12\%, compared to $\approx$4\% with the vanilla GPT-2 model and in most cases even slightly less than that with the early-stopped SLR models). This leads to overall worse perplexity.

\subsection{Differences between Formalisms}

Comparing across rows in \cref{tab:main_results}, we find a considerable performance spread.
The general trend, which is relatively consistent in all metrics,\footnote{The multitude of metrics might thus seem redundant. But since each measurement emphasizes different properties of model performance, we consider it a a very interesting result (and, potentially, a success of our modeling technique and experimental setup) to achieve this broad consistency.} is indicated by the order of rows, with UD having the smallest (though still respectable) improvement over the baseline, and PTG and EDS the largest.

\begin{table}[t]
    \centering\small
    \setlength{\tabcolsep}{3.7pt}
    \begin{tabular}{c | ccc | c}
            & \textbf{Dep}  & & \textbf{Const} & \textbf{Avg} \\\midrule
  \textbf{Syn}   & 32.7 \textcolor{gray}{\scriptsize \phantom{$\pm 9.7$}} (1) &  *** & 29.4 \textcolor{gray}{\scriptsize $\pm 0.6$} (2) & 30.5 \textcolor{gray}{\scriptsize $\pm 2.0$} (3) \\
                 &                                ***                         &   &       ***             &   $-$                  \\
  \textbf{Sem} & 31.0 \textcolor{gray}{\scriptsize $\pm 0.5$} (2) &      ***       & \textbf{25.7} \textcolor{gray}{\scriptsize $\pm 1.5$} (2) & 28.4 \textcolor{gray}{\scriptsize $\pm 3.2$} (4) \\\midrule
       \textbf{Avg} & 31.6 \textcolor{gray}{\scriptsize $\pm 1.0$} (3) &    **      &  27.6 \textcolor{gray}{\scriptsize $\pm 2.3$} (4) & 29.3 \textcolor{gray}{\scriptsize $\pm 2.8$} (7) \\
    \end{tabular}
    \caption{Model perplexity (lower is better) summarized in terms of two SLR dimensions: Scope (syntax vs.~semantics) and structure (dependency vs.~constituency). $\mu$ \textcolor{gray}{\scriptsize $\pm$ $\sigma$} ($n$) over frameworks per condition. Statistical significance of the difference between the two closest SLRs of each pair of conditions is reported as ***$p<.0001$ / **$p<.001$ / *$p<.005$ / $-$not significant (approximate randomization test with $R=$10,000 shuffles).}
    \label{tab:slr-dimensions-results}
\end{table}

Interestingly, there are two marked separations: a primary one between dependency and constituency formalisms, and a secondary one between syntactic (i.e., more surface-oriented) and semantic (more abstract) formalisms. This is summarized in \cref{tab:slr-dimensions-results}.
A limiting factor for dependency representations in the incremental LM setting is that relations between the target token and subsequent tokens are entirely ignored, whereas constituency graphs can back off to higher-level structures. Further, the syntactic graphs we use are always trees, so they never populate the coparent capacity in the slices.
\textit{Semantic constituency} representations, with their abstract and meaning-oriented labeling and structure schemes, jump out as being especially predictive of the underlying text, as compared to both syntax and shallow semantics.

We note that the function-enhanced PTB label set has a slight advantage over the basic phrase-structure labels; and that, among the two closely related pairs of formalisms (DM\slash EDS and PSD\slash PTG, which each are dependency and constituency versions converted from the same underlying grammars), the constituency versions always work better than the dependency versions in our setting. There is, however, no consistent ranking between DM\slash EDS on one hand and PSD\slash PTG on the other. In terms of perplexity, EDS works better than PTG, and PSD better than DM, but these differences are not significant for accuracy.

\subsection{Differences between Word Classes}\label{sec:pos}

To better understand where particular strengths and weaknesses of the baseline LM and linguistically enhanced models lie, we analyze subsets of tokens by part-of-speech (POS) tag (\cref{tab:pos_results}, see \cref{app:pos} for more details).
Across all models there is a clear and expected separation between rather predictable function words,
more perplexing content words, and numbers, punctuation, and miscellaneous tokens somewhere in the middle.

Average perplexity of the tested SLR models is better than baseline GPT-2 in all POS classes but one.
The one exception is the noun class, where both the SLR macro-average and UD in particular do not raise performance.
Only EDS and DM show perplexity improvements on nouns; PTB even has a noticeable negative impact.
We conjecture that this may have to do with relatively deep NP nesting in PTB (compared to the other formalisms), such that the current slicing hyperparameters (relative types and capacities) are too strict and hide informative signals like modifiers and verb attachment.

Some formalisms seem to be particularly well-suited for the prediction of certain POS: UD for verbs; PTB and PTG for adpositions and subordinating conjunctions; EDS for pronouns, determiners, and numbers; PTG, PSD, and EDS for coordinating conjunctions.
The advantage of EDS and DM on nouns, pronouns, determiners, and numbers can likely be attributed to their explicit representation of variable binding\slash quantification.
Similarly, PTG and PSD have detailed categories for coordination, distinguishing, e.g., con- and disjunction.

For nouns and modifiers, the spread across formalisms is particularly wide, which suggests that SLRs diverge quite a bit on these types of words (e.g., whether adjectives and certain nouns can count as predicates) and that this diversity has a strong effect on utility for language modeling.

\subsection{Model Ablations}\label{sec:abl}

The linguistically enriched models consist of a substantial number of newly learned parameters---around 50--60M each, an additional $\approx$50\% the size of vanilla GPT-2. Although model size does not seem to be correlated with performance among the SLR-enriched models, it could still be that the additional capacity allows the models to store more information about the words' distributions than the baseline GPT-2 model, without ever truly using the concrete linguistic structures.

We check this by randomly shuffling ($\neovsearrow$) two core graph properties: (i) the assignment of edge \textit{labels}, and (ii) the \textit{anchoring} mapping between graph nodes and word tokens in each graph.  
If the models are largely independent of the correct label and structure assignments, 
these changes should have a very small effect on performance \citep{dubossarsky-etal-2018-coming,hewitt-liang-2019-designing}.

\begin{table}[t]
    \centering\small
    \setlength{\tabcolsep}{2.2pt}
    \begin{tabular}{c| lrr rrrr}
\multicolumn{1}{c}{} && \multirow[b]{2}{*}{\begin{tabular}{c}
            Eval \\
            \textbf{Toks}
        \end{tabular}} &
        \multirow[b]{2}{*}{\begin{tabular}{c}
            Train \\
            \textbf{Vocab}
        \end{tabular}} \hspace*{-5pt} & 
        \multicolumn{4}{c}{{Perplexity}\textdownarrow} \\\cmidrule{5-8}
\multicolumn{1}{c}{} & \textbf{POS} & & & \multicolumn{1}{c}{\textbf{GPT-2}} & \multicolumn{1}{c}{\textbf{UD}} & \multicolumn{1}{c}{\textbf{EDS}} & \multicolumn{1}{c}{\textbf{SLR Avg}} \\\toprule
\multicolumn{1}{c}{} & All & 22,596 & 27,344 & 45.9 & 32.7	  &		\textbf{24.7}	  &		29.3	 \textcolor{gray}{\scriptsize $\pm \phantom{0}2.8$}  \\\midrule
\multirow{3}{*}{\rotatebox{90}{content}} & noun & 7,731 & 18,435 & 142.5 & 122.0	  &		\textbf{98.0}	 &		122.6	\textcolor{gray}{\scriptsize $\pm 13.9$}  \\
 & verb & 2,639 & 7,100 & 128.8 & \textbf{80.4}	  &		85.9	 &		84.9	\textcolor{gray}{\scriptsize $\pm \phantom{0}4.5$}  \\
 & mod & 2,235 & 6,292 & 228.7 & 158.8	  &		\textbf{98.6}	 &		124.4 	\textcolor{gray}{\scriptsize $\pm 22.6$}  \\
\multicolumn{1}{c}{\vspace*{-5pt}} \\
\multirow{7}{*}{\rotatebox{90}{function}} & aux & 582 & 95 & 17.6 & 11.1	  &		\textbf{5.9}	 &		9.1	\textcolor{gray}{\scriptsize $\pm \phantom{0}2.1$} \\
 & adp & 1,957 & 232 & 10.1 & 7.3		  &		5.5		  &		\textbf{5.3}		 \textcolor{gray}{\scriptsize $\pm \phantom{0}1.6$}  \\
 & part & 645 & 27 & 3.7 & 2.0		  &		\textbf{1.6}		  &		1.9		 \textcolor{gray}{\scriptsize $\pm \phantom{0}0.3$}  \\
 & sconj & 268 & 96 & 15.4 & 12.3		  &		6.8		  &		\textbf{6.8}		 \textcolor{gray}{\scriptsize $\pm \phantom{0}3.9$}  \\
 & cconj & 548 & 35 & 13.0 & 7.4		  &		\textbf{1.9}		  &		4.1		 \textcolor{gray}{\scriptsize $\pm \phantom{0}2.1$}  \\
 & det & 1,726 & 91 & 9.4 & 7.8	  &		\textbf{4.4}	 &		6.0	\textcolor{gray}{\scriptsize $\pm \phantom{0}1.3$}  \\
 & pron & 868 & 149 & 22.9 & 17.5	  &		\textbf{5.4}	 &		11.0	\textcolor{gray}{\scriptsize $\pm \phantom{0}4.0$}  \\
\multicolumn{1}{c}{\vspace*{-5pt}} \\
\multicolumn{1}{c}{} & num & 719 & 1,059 & 72.6 & 57.1	  &		\textbf{47.5}	 &		54.1	\textcolor{gray}{\scriptsize $\pm \phantom{0}4.6$}  \\
\multicolumn{1}{c}{} & punct & 2,527 & 68 & 4.9 & \textbf{2.3}	  &		2.7	 &		2.6	\textcolor{gray}{\scriptsize $\pm \phantom{0}0.3$}  \\
\multicolumn{1}{c}{} & misc & 151 & 183 & 7.0 & 4.6	  &		\textbf{4.0}	 &		4.5	\textcolor{gray}{\scriptsize $\pm \phantom{0}0.8$}  \\\bottomrule
    \end{tabular}
    \caption{Breakdown by Universal POS,\footnotemark{} in terms of PPL of domain-trained GPT-2, two exemplary SLR-combined models, and the macro-average $\pm$ stdev over all SLR-combined models. Best results (within the variance) in each row are \textbf{bolded}. We show token counts and observed vocabulary size for reference.
    }
    \label{tab:pos_results}
\end{table}

\footnotetext{\url{https://universaldependencies.org/u/pos/}}

\begin{table}[t]
    \centering\small
    \setlength{\tabcolsep}{3.8pt}
    \begin{tabular}{lc rrr}
         \multicolumn{1}{c}{\textbf{Ablation}} & \multicolumn{1}{c}{\textbf{Applied in}} & \multicolumn{1}{c}{\textbf{DM}} & \multicolumn{1}{c}{\textbf{PTB}} & \multicolumn{1}{c}{\textbf{SLR Avg}} \\\toprule
        \textit{Full} &  &         31.4 &         29.0 &     29.3 \textcolor{gray}{\scriptsize $\pm \phantom{0}2.8$} \\\midrule
$\neovsearrow$ Labels & testing &  $+$4.7 &  $+$73.9 &  $+$28.3 \textcolor{gray}{\scriptsize $\pm 28.3$} \\
$\neovsearrow$ Anchors & testing &  $+$34.8 &  $+$223.1 &  $+$106.0 \textcolor{gray}{\scriptsize $\pm 73.1$} \\
$\neovsearrow$ Both & testing &  $+$33.4 &  $+$207.4 &  $+$95.9 \textcolor{gray}{\scriptsize $\pm 68.9$} \\[3pt]
$\neovsearrow$ Labels & training &  $+$1.4 &  $+$9.0 &  $+$4.2 \textcolor{gray}{\scriptsize $\pm \phantom{0}3.3$} \\
$\neovsearrow$ Anchors & training &  $+$8.5 &  $+$17.5 &  $+$13.3 \textcolor{gray}{\scriptsize $\pm \phantom{0}4.6$} \\
$\neovsearrow$ Both & training &  $+$7.8 &  $+$18.3 &  $+$13.4 \textcolor{gray}{\scriptsize $\pm \phantom{0}5.0$} \\[3pt]
$\neovsearrow$ Labels & both &  $+$1.3 &  $+$9.3 &  $+$4.3 \textcolor{gray}{\scriptsize $\pm \phantom{0}3.4$} \\
$\neovsearrow$ Anchors & both &  $+$7.9 &  $+$17.5 &  $+$13.5 \textcolor{gray}{\scriptsize $\pm \phantom{0}5.0$} \\
$\neovsearrow$ Both & both &  $+$7.3 &  $+$18.1 &  $+$13.6 \textcolor{gray}{\scriptsize $\pm \phantom{0}5.2$} \\[3pt]
        $-$ SLR & both &        $+$14.5 &      $+$16.9 &  $+$16.6 \textcolor{gray}{\scriptsize $\pm \phantom{0}2.8$}  \\
        \bottomrule
    \end{tabular}
    \caption{Ablations measured in $\Delta$PPL for two exemplary SLR-combined models and the macro-average $\pm$ stdev over all SLR-combined models. 
    \textit{Full} and $-$SLR correspond, respectively, to \cref{tab:main_results}'s rows 4 (DM) \slash\ 7 (PTB-fxn) and row 2 (GPT-2 $+$Domain).
    }
    \label{tab:abl_results}
\end{table}

But on the contrary, we find that performance worsens considerably in the ablated settings compared to the full combined models of each formalism (\cref{tab:abl_results}, see \cref{app:abl} for more details).
This confirms that the models really do acquire---and are quite sensitive to---the graph-encoded linguistic signals, relying to a large part on this new information in making their predictions.

Shuffling only edge labels while leaving the rest of the graphs unchanged has a smaller effect than changing how tokens are anchored in the graph structure. 
This suggests that the linguistic graphs' entire structural arrangement of labels and attention-like selection of context words play a crucial role---more so than knowing the type of each individual (correctly attached) grammatical relations.
Note that the $\neovsearrow$ Anchors setting, too, changes which edge labels are used in the prediction of a given token, resulting in a smaller difference between $\neovsearrow$ Anchors and $\neovsearrow$ Both.

If a model has learned to rely on correct labels and structure during training, then perturbing these properties at test time has a highly adverse effect, confusing the model and leading to a drastic decrease in performance---even worse than not consulting SLR graphs at all!
Given previous findings that syntactic structure is to some extent already learned in pretraining \citep{linzen-etal-2016-assessing,tenney2018what}, we conjecture that this representational capacity gets offloaded to the graphs at training time, and thus test-time permutations fool the PTB model to a much greater extent than DM.

As expected, exposing models to shuffled graphs at training time renders the additional model parameters practically neutral, resulting in similar perplexity as the base LM.
In this case, it also does not matter whether test-time graphs are correct or random (training vs.~both in column 2)---either way, the model learns to mostly disregard the random structure as noise.

\subsection{Comparison with R-GCN Encoding}\label{sec:gnn-baseline}

As an additional strong baseline, we compare our concatenation-based slice vector encoding to a graph neural network from the literature.
We choose relational graph-convolutional networks \citep[\mbox{R-GCN};][]{schlichtkrull-etal-2018-rgcn,kipf_semi-supervised_2017} as a suitable representative of this type of model, which has been used successfully by \citet{wu-2021-infusing} to encode DM graphs.

\begin{table}[t]
    \centering\small
    \setlength{\tabcolsep}{4.5pt}
    \begin{tabular}{l cr cc}
    & \multicolumn{2}{c}{Training Efficiency} & \multicolumn{2}{c}{LM Quality}\\
    \cmidrule(r){2-3}\cmidrule(l){4-5}
        \multicolumn{1}{l}{\textbf{Model}} & \multicolumn{1}{c}{$\delta$\textbf{Speed} \textuparrow} & \multicolumn{1}{c}{$\Delta$\textbf{Size} \textdownarrow} & \multicolumn{1}{c}{$\Delta$\textbf{PPL} \textdownarrow} & \multicolumn{1}{c}{$\Delta$\textbf{Acc} \textuparrow} \\\toprule
        UD & $-$50\% & $-$1.9M & $+$2.9 \textcolor{gray}{\scriptsize $\pm .08$} & $-$0.4 \textcolor{gray}{\scriptsize $\pm .02$} \\[3pt]
        DM & $-$47\% & $+$2.5M & $+$1.6 \textcolor{gray}{\scriptsize $\pm .35$} &	$+$0.1 \textcolor{gray}{\scriptsize $\pm .14$} \\
        PSD & $-$56\% & $+$9.1M &  $+$3.6	 \textcolor{gray}{\scriptsize $\pm .23$} & $-$0.9 \textcolor{gray}{\scriptsize $\pm .15$} \\[3pt]
        PTB-phr & $-$43\% & $-$1.9M & $+$6.8 \textcolor{gray}{\scriptsize $\pm .39$} &  $-$1.7 \textcolor{gray}{\scriptsize $\pm .11$} \\
        PTB-fxn & $-$86\% & $+$107.7M & $+$10.5  \textcolor{gray}{\scriptsize $\pm .21$} & $-$2.7 \textcolor{gray}{\scriptsize $\pm .15$} \\[3pt]
        PTG & $-$53\% & $+$5.9M & $+$6.2 \textcolor{gray}{\scriptsize $\pm .22$} &	$-$3.5 \textcolor{gray}{\scriptsize $\pm .03$} \\
        EDS & $-$47\% &  $-$8.5M & $-$0.2 \textcolor{gray}{\scriptsize $\pm .07$} &	$+$0.1 \textcolor{gray}{\scriptsize $\pm .03$}\\  
        \bottomrule
    \end{tabular}
    \caption{Performance differences between R-GCN slice encoder baseline and our concatenation-based encoder (\cref{tab:main_results}). Relative differences ($\delta$) for speed in sentences per second; absolute differences ($\Delta$) otherwise. Means $\pm$ stdev over 2 runs without early stopping.}
    \label{tab:rgcn_results}
\end{table}

Results are shown in \cref{tab:rgcn_results}.
Contrasting with \cref{tab:main_results}, there is a big difference in training speed: our simple encoder is on average roughly twice as fast as the computation-heavy alternative, whose time and space complexity is dominated by the number of labels.\footnote{And this is a very optimistic estimate of R-GCN training speed in practice; see \cref{app:rgcn-speed}.}

We observe at best similar LM quality as with our concatenation method (EDS and DM), but for most formalisms performance degrades.
We follow \citeauthor{schlichtkrull-etal-2018-rgcn} and \citeauthor{wu-2021-infusing} in using 2 R-GCN layers with basis matrix regularization.
Possible disadvantages of this for encoding linguistic graphs are the fixed path length (2 layers exclude parent's siblings; but 3 layers would include a lot of irrelevant information) and that many of the trained parameters are shared between different relations. In contrast, our concatenation encoding forces the MLP input layer to learn distinct parameters for each structural relative type and edge label.

\section{Discussion}\label{sec:discussion}

\subsection{Related Work}\label{sec:rel-work}

Researchers have long been interested in scaffolding sequential language models with linguistic-structure-based inductive biases.
\textit{Syntactic language modeling} dates back to the pre-neural era, when \citet{pauls-klein-2012-large} and \citet{gubbins-vlachos-2013-dependency} generalized Markov assumptions from word \textit{n}-grams to syntactic subtrees.
These ideas have since been adapted to recurrent neural network (RNN) LMs \citep{mirowski-vlachos-2015-dependency} and expanded on \citep{dyer-16-recurrent,choe-charniak-2016-parsing,shen2018neural,shen_ordered_2019}.
\citet{ek-etal-2019-language} condition RNN-LMs on predicted syntactic and semantic (unstructured) tags, interestingly finding less or sometimes no benefit, especially on the semantic side. 
They hypothesize this might be due to tagging errors---an issue our oracle setup avoids.

In the era of attention-based neural modeling of language dominated by pretrained Transformers, models are often finetuned for and evaluated on specific NLP tasks---like semantic role labeling, machine translation, natural language inference, graph-to-text generation, or the GLUE benchmark \citep{wang-etal-2018-glue}---rather than language modeling in its own right, which makes it difficult to compare them directly to our findings.
There have been two main directions:
One group of approaches continues the old syntactic language modeling tradition by incrementally generating words and SLRs with either joint \citep{peng-etal-2019-palm,qian-etal-2021-structural,sartran_transformer_2022}
or iteratively-coupled LM and parser models \citep{choshen2021transition}.
The second group assumes parsed input sentences, which are then used to guide the model, e.g.~by directly optimizing Transformers' attention weights to reflect linguistic graph structures \citep{strubell-etal-2018-linguistically,bai-etal-2021-syntax,slobodkin2021semantics}.
Rather than controlling the existing sequential attention, \citet{hajdik-etal-2019-neural} process serialized graphs directly with a sequence-to-sequence model, and \citet{wu-2021-infusing} extend a pretrained Transformer with an additional graph encoder.
Notably, \citet{wu-2021-infusing} and \citet{slobodkin2021semantics} experiment with a few different semantic and syntactic SLRs, while all other studies we have looked at are limited to either syntax or very shallow semantics.

Another relevant line of work employs \textit{probing tasks} in investigating to what extent grammar and meaning are already encoded in neural language models trained predominantly on raw text with little to no linguistic supervision \citep[][\textit{inter alia}]{linzen-etal-2016-assessing,tenney-etal-2019-bert,tenney2018what,hewitt-19-probe,liu2019linguistic,kim-etal-2019-probing,wu-etal-2020-perturbed,Geiger:Lu-etal:2021-abstractions}.
Among the probing literature, the works of \citet{kuznetsov-gurevych-2020-matter} and \citet{kulmizev-etal-2020-neural} are noteworthy in that they investigate subtle differences between different (versions of) frameworks roughly covering the same representational scope, namely, semantic roles and syntactic dependencies, respectively.

Orthogonal approaches to \textit{comparing SLR designs} have involved measuring how well different frameworks complement each other for joint parsing or can be merged or converted into one another \citep{prange-etal-2019-uccasnacs,hershcovich-etal-2020-comparison}.

\subsection{Limitations and Future Work}

While the use of oracle graphs has both theoretical advantages (measuring an upper bound without needing to account for potential errors or uncertainties) and practical ones (saving the computational overhead from training and running a parser), ground-truth SLR graphs are a very limited resource and generally assumed to only be available at training time.
There is no guarantee our results translate to the non-oracle setting. For instance, it could be that the most helpful abstract semantic information
is also the hardest to predict.
And despite segmenting the existing sentence-level graph into token-level slices, the human annotator who created the graph in the first place has seen and analyzed the whole sentence, thus already resolving crucial ambiguities and simplifying the task based on knowledge `from the future'.
In subsequent work, we plan to \textit{parse} graph slices incrementally, which will both relax the \textit{conditional} modeling assumption into a more broadly interpretable \textit{joint} model and enable test-time use of the full system on datasets without linguistic annotations.

We also only test formalisms that are explicitly anchored in linguistic units, roughly corresponding to LM (sub-)word tokens. This prevents us from applying the same paradigm to some other widely-used \textit{unanchored} formalisms like AMR \citep{banarescu2013abstract} without some changes to the setup.

\subsection{Broader Impact}

Our experiments yield evidence which---at least in the case of encoding contexts for next-word prediction---supports the thesis of \citet{bender-koller-2020-climbing}, \citet{trott-etal-2020-construing}, and others that linguistic \textit{meaning} goes beyond \textit{form}.
Computational models of language that exclusively learn from even very large amounts of raw text are thus generally expected to hit a ceiling\footnote{See also \citet{merrill-etal-2021-provable} for formal proofs.}
which can only be overcome with access to higher-level structures and mechanisms of understanding.

It further seems to matter in which manner and shape linguistic graph structure is drawn. Assuming a perfect incremental parser, deeper structure and semantic categorization seems to be particularly beneficial for integration with a standard language model. This is in line with previous findings by, e.g., \citet{tenney2018what} that while pretrained LMs tend to encode shallow syntactic structure, abstract relations are more difficult to probe for.

We thus see a promising research direction in moving towards linguistic scaffolding of language models with representations that are \textit{more complex} than tags or dependencies and that capture \textit{meaningful relations} beyond surface structure.

\section{Conclusion}

We have presented evidence that symbolic linguistic representations of various frameworks have the potential to aid a pretrained incremental Transformer in task-neutral next-word prediction. To this end, we have proposed a framework-agnostic neural encoding scheme for linguistic graphs and
applied it to an English dataset jointly annotated with 7 different formalisms. The results highlight the importance of appreciating complex linguistic structure and handling its computational representation with nuance.

\section*{Acknowledgements}

We would like to thank Katrin Erk and Chris Dyer; members of the Georgetown NERT/GUCL and HKU NLP labs; the organizers, reviewers, and audience of MASC-SLL 2022; as well as the anonymous ARR reviewers for their extremely insightful feedback and suggestions.

\bibliographystyle{acl_natbib}
\bibliography{diss}

\clearpage

\appendix

\section{Additional Modeling Details}

\subsection{Selecting Anchor Nodes}\label{app:anchor}

In case there are multiple anchoring options (see, e.g., EDS nodes 0 vs.~1 for first token in \cref{fig:slice}), we use the following tie-breaker heuristics: Select the anchor node with the most parents and children; if still a tie, select the anchor with the highest node ID (tends to be hierarchically lower, i.e., vertically closer to the token anchor).

\subsection{Relative Types}\label{app:rels}

$REL = \langle P, B, O, T, C, R\rangle$, namely, parents $P_{i}$, siblings $B_{i, p}$, grandparents $O_{i, p}$, aunts $T_{i, p}$ (all indexed by parent $p$), children $C_i$, and coparents $R_{i, c}$ (indexed by child $c$).
This is the anchor node's Markov blanket, plus siblings, grandparents, and aunts. We chose this set of relations based on general notions of linguistic hierarchy (predicate-argument, head-dependent) and preliminary experiments, but without tuning for specific formalisms.
Precise definitions are given in \cref{tab:relcap}.
Relative nodes are permanently associated with the label of the edge that got them selected.

\subsection{Representing Tokens and Labels}\label{app:enc}

We use \mbox{GPT-2's} pretrained global embeddings (from the lowest layer, before any local contextualization) to obtain embeddings for relative token anchors in the slice-vector. When a token anchor in a linguistic graph consists of multiple BBPE tokens, we average their embeddings.
We reuse the transpose of the same embedding matrix again to project the last hidden state of the token-emission MLP into the vocabulary.

SLR edge labels are encoded as one-hot vectors in the slice vectors, which lowers the potential for unnecessary random initialization variance of from-scratch embeddings.

\subsection{Distinguishing Dependencies from Constituencies}\label{app:const-dep}

\begin{figure}[t]
    \centering
    \input{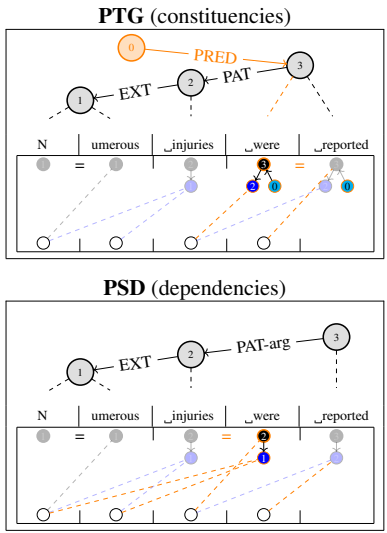}
    \caption{Example of subtle differences in constituency (PTG) and dependency (PSD) versions of the same underlying formalism, the Prague Functional Description. PTG has an abstract PRED node as well as a multiword anchor where PSD does not, which results in diverging slice representations for the last two tokens.}
    \label{fig:const-dep}
\end{figure}

While this distinction---as defined in \cref{sec:formalisms} in terms of the anchoring mapping between graph nodes and word tokens---can be subtle for individual sentences, it nonetheless affects slice encoding.
In PSD, for example, auxiliaries are unanchored, whereas in PTG they are grouped with their main predicate (\cref{fig:const-dep}).

\subsection{Model Hyperparameters}\label{app:hyper}

We report our model and training hyperparameters in \cref{tab:hyper}. We did \textit{not} perform explicit hyperparameter tuning, besides some manual testing early in development on a subset of the MRP shared task data. Those data are annotated with SLR frameworks other than the ones we compare here, and we ended up excluding them from our experiments for lack of overlap with most of the other frameworks' annotations.

\subsection{Efficient Batching for R-GCN}\label{app:rgcn-speed}

In our incremental setting we need to apply the \mbox{R-GCN} to each token-level slice, which would lead to multiple days\footnote{Projected timeline based on a few iterations, which is confirmed by \citet{yang_deng_2020_incremental}.} of training for each model if done naively.
We achieve a considerable speedup by exploiting the oracle graphs at training and evaluation time to pre-compute slices and running the R-GCN only once per sentence batch.

\begin{table}[t]
    \centering\small
    \begin{tabular}{lllc}
        \textbf{rel} & \textbf{Name} & \textbf{Definition} & \textbf{$\gamma$} \\\toprule
        $P_{i}$ & parent & $\{v : (v, a_i) \in E\}$ & 2\\
        $B_{i, p}$ & sibling & $\{v : (p, v) \in E\}\ \forall p \in P_{i}$ & 2 \\
        $O_{i, p}$ & grandparent & $\{v : (v, p) \in E\}\ \forall p \in P_{i}$ & 1 \\
        $T_{i, p}$ & aunt & $\{v : (o, v) \in E \wedge o \in O_{i, p}\}$ & 2 \\ 
        & & $\forall p \in P_{i}$ \\\midrule
        $C_{i}$ & child & $\{v : (a_i, v) \in E\}$ & 2 \\
        $R_{i, c}$ & coparent & $\{n : (v, c) \in E\}\ \forall c \in C_{i}$ & 1 \\  
        \bottomrule
    \end{tabular}
    \caption{Relative types and capacities.}
    \label{tab:relcap}
\end{table}

\begin{table}[t]
    \centering\small
    \begin{tabular}{lrc}
    \toprule
    \multicolumn{2}{l}{\textbf{GPT-2}} \\\midrule
        Embedding dim & 768 \\
        Vocabulary & 50,257 \\
        Activation & GELU \\
        Dropout & 0.1 \\
        Learning rate & 1e-6\\\midrule
    \multicolumn{2}{l}{\textbf{MLP}} \\\midrule
        Input dim & \multicolumn{2}{c}{$16*|L| + 17*768$\footnotemark{}} \\
        Layers & 2 \\
        Hidden dims & 1,024; 768 \\
        Activation & ReLU \\
        Dropout & 0.2 \\
        Learning rate & 1e-4\\\midrule
    \multicolumn{2}{l}{\textbf{R-GCN}} \\\midrule
        Input dim &  768 \\
        Layers &  2 \\
        Hidden dims &  768; 768 \\
        Activation & ReLU \\
        Basis matrices & \multicolumn{2}{c}{$\lfloor 0.1*|L^*|\rfloor$\footnotemark{}}  \\
        Learning rate & 1e-4\\\midrule
    \multicolumn{2}{l}{\textbf{Other training settings}} \\\midrule
        Epochs & 10 \\
        Batch size & 8\\\bottomrule
    \end{tabular}
    \caption{Model and training hyperparameters}
    \label{tab:hyper}
\end{table}

\addtocounter{footnote}{-1}

\footnotetext{For label set $L$. The factor 16 arises from the capacities chosen (\cref{tab:relcap}), and the extra embedding allocation is for averaged preceding unanalyzable/within-anchor tokens.}
\stepcounter{footnote}
\footnotetext{For bidirectional label set $L^*$, which is twice as big as $L$.}

\section{Data Preprocessing}\label{app:preproc}

\subsection{Sentence Filtering}\label{app:sent-filter}

To establish a common ground for comparison, we take the intersection of sentences occurring in the annotated datasets of \textit{all} linguistic formalisms.

In a first step, we discard two sentences whose linguistic graph in at least one formalism is empty\footnote{The sentence ``It is.'' in DM and a `sentence' consisting of the @-symbol in PTG.}. 
We then select only those 35,513 train-dev \slash\ 1,401 eval sentences that appear in both the MRP 2019 and 2020 datasets (the 2019 corpus contains 143\slash 1,958 more in train-dev\slash eval).\footnote{`train-dev' refers to the data split that was used as training data in both the MRP and 2019 tasks, and which we split 90\%\slash 10\% into our training and development data. `eval' refers to the data that was used as evaluation data in MRP 2019 and as development data in MRP 2020, and which we evaluate our models on.}
Next, we take the intersection of these sentences and OntoNotes 5.0, which contains the gold PTB syntax annotations. 26,719\slash 929 sentences remain in the train-dev\slash eval set.
The MRP graph format operates on raw-text character offsets, while PTB and UD trees operate on word tokens. We are able reconstruct offset-based text anchors for PTB and UD from the raw text strings used in the MRP data for all but 394 train-dev \slash\ 8 eval sentences, which leaves us with the final 26,325 train-dev and 921 eval sentences.

In a few cases, where the linguistic graph has no edges, we add an artificial edge with a dummy label.

\subsection{Tokenization}\label{app:tok}

We follow the sentence segmentation of the Penn Treebank corpus. 
Within sentences, we obtain token boundaries from GPT-2's pretrained byte-level byte-pair encoding (BBPE) tokenizer. The BBPE tokens are then aligned with the formalism-dependent SLR node anchors via raw-text character offsets.
Tokens that are continuations of multiword anchors in the graph (`\textvisiblespace reported' in PTG, \cref{fig:slice}); subword tokens of a single graph anchor (`\mbox{N-umerous}'); or are unanchored in the graph (`\textvisiblespace were' in EDS), are treated as \textit{unanalyzable}, i.e., their slice consists of a copy of the preceding token's slice, plus the preceding within-anchor tokens.

\subsection{UD Conversion}\label{app:ud}

Quasi-gold UD 2.0 trees are obtained from the UD converter released with the Java Stanford Parser v4.2.0 (\url{https://nlp.stanford.edu/software/lex-parser.html}) on the PTB trees.

\subsection{PTB Labels}\label{app:ptb}

By convention, phrasal and functional labels in PTB are node labels. To match the labeled-edges-unlabeled-nodes format of the other formalisms, we losslessly convert them to edge labels (namely, on each node's single incoming edge), discarding the preterminal nodes' POS labels. In preliminary experiments we saw that including the POS tags is much more beneficial than phrase structure only; but since we do not include word-level tags in any of the other conditions, this would be an unfair comparison. We focus here on sentence-level structure and leave studies of word-level tags to future work.

\subsection{Data Splits}\label{app:splits}

We split the corpus into training\slash development and evaluation data following the MRP task setup. Specifically, we evaluate on the data split that was used as evaluation data in MRP 2019 and as development data in 2020, as only for this data gold annotations in all formalisms have been released. We do not perform empirical hyperparameter tuning. In early development, a small subset of the data was used.

\begin{table*}[p]
    \centering\small
    \setlength{\tabcolsep}{4.5pt}
    \begin{tabular}{l ccc ccH  c c c c c}
    & & & & \multicolumn{2}{c}{Training Efficiency} & & \multicolumn{5}{c}{Language Model Quality}\\
    \cmidrule{5-6}\cmidrule{8-12}
        \multicolumn{1}{l}{\textbf{Model}} & \multicolumn{2}{c}{\textbf{Scope/Struct}} & \multicolumn{1}{c}{\textbf{\#Labels}} & 
        \multicolumn{1}{c}{\textbf{Speed} \textuparrow} & \multicolumn{1}{c}{\textbf{Size} \textdownarrow} & \multicolumn{1}{H}{\textbf{|Slice|}} & \multicolumn{1}{c}{\textbf{PPL} \textdownarrow} & \multicolumn{1}{c}{\textbf{H} [nats] \textdownarrow} & \multicolumn{1}{c}{\textbf{Acc} [\%] \textuparrow} & \multicolumn{1}{c}{\textbf{Conf} [\%] \textuparrow} & \multicolumn{1}{c}{\textbf{MRR} \textuparrow}  \\\toprule
        GPT-2 & & & & -- & \multirow{2}{*}{124.4M} & -- & 59.3	\phantom{\scriptsize $\pm .00$} & 4.09 \phantom{\scriptsize $\pm .000$} & 30.0 \phantom{\scriptsize $\pm .00$} & \textbf{31.2} \phantom{\scriptsize $\pm .00$} & .403	\phantom{\scriptsize $\pm 0.0$e-3} \\
        + Domain & & & & 15 & & -- & 45.8	\textcolor{gray}{\scriptsize $\pm .03$} &	3.61	\textcolor{gray}{\scriptsize $\pm .002$} &	33.4	\textcolor{gray}{\scriptsize $\pm .05$} &	35.3	\textcolor{gray}{\scriptsize $\pm .02$} &	.436	 \textcolor{gray}{\scriptsize $\pm \phantom{0}.3$e-3} \\\midrule
        + UD & syn & dep & \phantom{0}39 & 14 & +54.1M & 13.7K & 35.2 \textcolor{gray}{\scriptsize $\pm .24$} & 3.09 \textcolor{gray}{\scriptsize $\pm .014$} & 39.2 \textcolor{gray}{\scriptsize $\pm .11$} & 42.3 \textcolor{gray}{\scriptsize $\pm .18$} & .488 \textcolor{gray}{\scriptsize $\pm \phantom{0}.8$e-3} \\[3pt]
        + DM & sem & dep & \phantom{0}59 & 15 & +54.4M & 14.0K & 34.2 \textcolor{gray}{\scriptsize $\pm .32$} & 3.05 \textcolor{gray}{\scriptsize $\pm .019$} & 38.8 \textcolor{gray}{\scriptsize $\pm .15$} & 42.5 \textcolor{gray}{\scriptsize $\pm .26$} & .490 \textcolor{gray}{\scriptsize $\pm 1.0$e-3}  \\[1pt]
        + PSD & sem & dep & \phantom{0}90 & 16 & +54.9M & 14.5K & 34.1 \textcolor{gray}{\scriptsize $\pm .43$} & 2.96 \textcolor{gray}{\scriptsize $\pm .014$} & 39.2 \textcolor{gray}{\scriptsize $\pm .17$} & 44.0 \textcolor{gray}{\scriptsize $\pm .17$} & .491 \textcolor{gray}{\scriptsize $\pm 1.4$e-3} \\[3pt]
        + PTB-phr & syn & const & \phantom{0}38 & 14 & +54.1M & 13.7K & 33.5 \textcolor{gray}{\scriptsize $\pm .30$} & 2.97 \textcolor{gray}{\scriptsize $\pm .026$} & 40.3 \textcolor{gray}{\scriptsize $\pm .09$} & 43.9 \textcolor{gray}{\scriptsize $\pm .34$} & .500 \textcolor{gray}{\scriptsize $\pm \phantom{0}.6$e-3}  \\[1pt]
        + PTB-fxn & syn & const & 537 & 14 & +62.7M & 22.1K & 32.4 \textcolor{gray}{\scriptsize $\pm .37$} & 2.92 \textcolor{gray}{\scriptsize $\pm .030$} & 41.1 \textcolor{gray}{\scriptsize $\pm .18$} & 44.8 \textcolor{gray}{\scriptsize $\pm .36$} & .507 \textcolor{gray}{\scriptsize $\pm 1.3$e-3}  \\[3pt]
        + PTG & sem & const & \phantom{0}72 & 15 & +54.6M & 14.3K & 29.6 \textcolor{gray}{\scriptsize $\pm .20$} & \textbf{2.68} \textcolor{gray}{\scriptsize $\pm .028$} & \textbf{43.4} \textcolor{gray}{\scriptsize $\pm .08$} & 48.8 \textcolor{gray}{\scriptsize $\pm .32$} & .524 \textcolor{gray}{\scriptsize $\pm \phantom{0}.5$e-3}  \\[1pt]
        + EDS & sem & const & \phantom{0}10 & 15 & +53.6M & 13.3K & \textbf{26.6} \textcolor{gray}{\scriptsize $\pm .09$} & 2.78 \textcolor{gray}{\scriptsize $\pm .024$} & 43.1 \textcolor{gray}{\scriptsize $\pm .10$}  & 46.6 \textcolor{gray}{\scriptsize $\pm .24$} & \textbf{.527} \textcolor{gray}{\scriptsize $\pm \phantom{0}.8$e-3}  \\\bottomrule
    \end{tabular}
    \caption{
    Main results \textbf{without early stopping}: performance of language models combined with 7 SLR formalisms of different scope, structure, and label set (each corresponding to a $P_{\textit{Ensemble}}$ in \cref{sec:p-slr}), compared to vanilla GPT-2 and a version of GPT-2 that has been domain-finetuned on the raw text of the SLR training corpus ($P_{LM}$). We report each quality metric as mean $\pm$ stdev over 5 random seeds. We also report model size in \#parameters and training speed in sentences per second as measures of efficiency. Best results in each column are \textbf{bolded}. For confidence, `best' means best-calibrated, i.e., the smallest relative difference to accuracy.
    }
    \label{tab:no-es-results}
\end{table*}

\section{Detailed Results}\label{app:results}\label{app:pos}\label{app:abl}

We report detailed results without early stopping (\cref{tab:no-es-results}), breakdowns by POS-class (\cref{tab:pos_results_ext,app:pos-input}), as well as ablation experiments (\cref{tab:abl_results_ext}) for all SLR formalisms.
In \cref{tab:pos_results,tab:pos_results_ext,fig:pos-input} we merge the POS tags \{NOUN, PROPN\} into `noun', \{ADJ, ADV\} into `mod', and \{INTJ, SYM, X\} into `misc'.

\subsection{Lexico-Semantic or Syntactic Knowledge?}\label{app:pos-input}

In \cref{sec:pos} we have found part-of-speech-specific patterns of model performance. But whenever, for a certain syntactic word class $S$, a formalism $A$ is more conducive to next-word prediction than a formalism $B$, it is not clear whether this is the case because the choices get narrowed down to $S$ itself or whether it is caused by either complementary or completely independent signals, perhaps at the lexical or semantic-structure levels.

We investigate this by rerunning the experiment with each token's UPOS tag as an additional input. If this is more or less the same information as is gained---to different extents---from the SLRs, then the results should be similar to before, and SLR-conditional differences should disappear.

\begin{figure*}[p]
    \centering
    \includegraphics[width=0.39\textwidth]{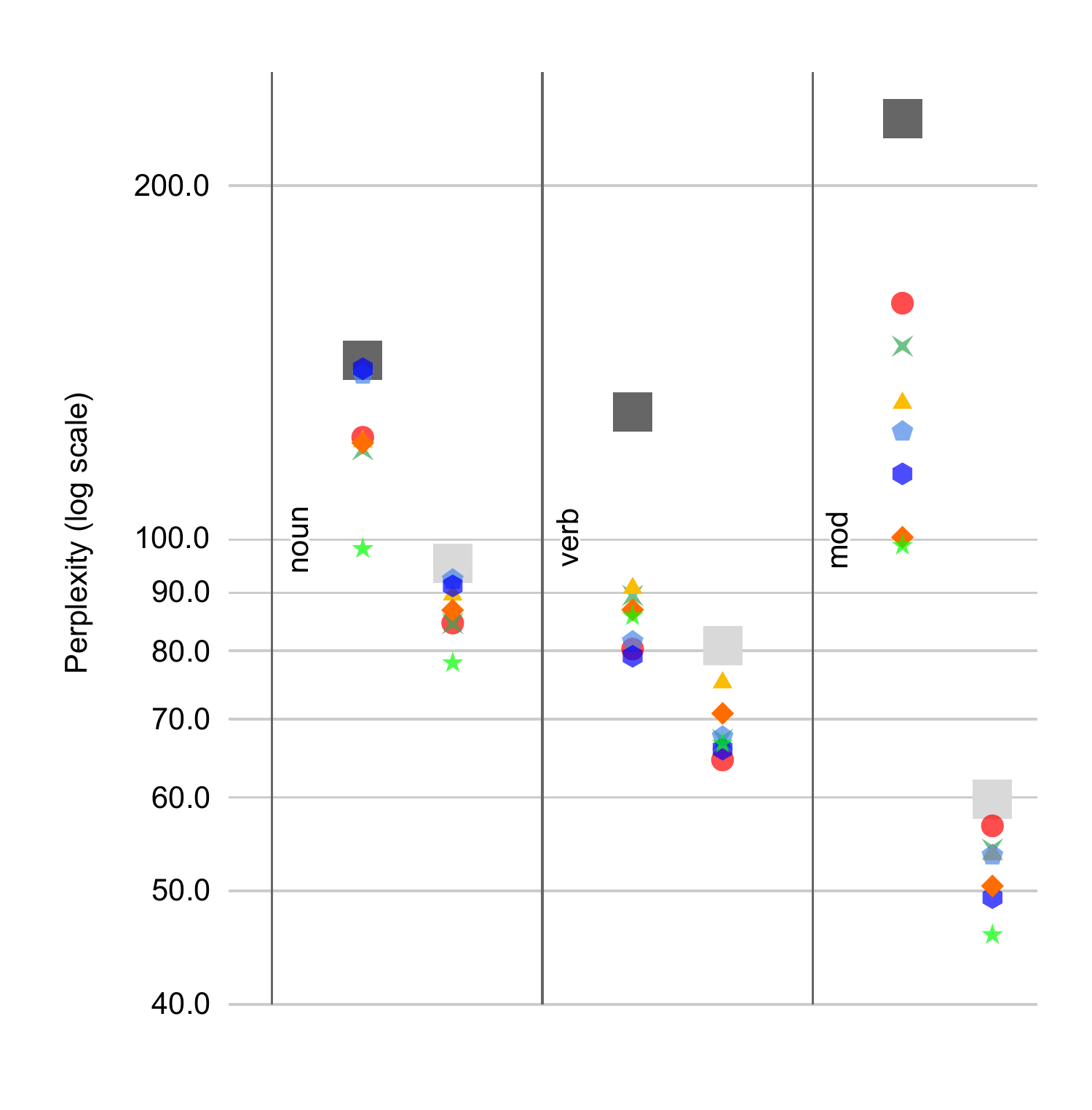}%
    \includegraphics[width=0.39\textwidth]{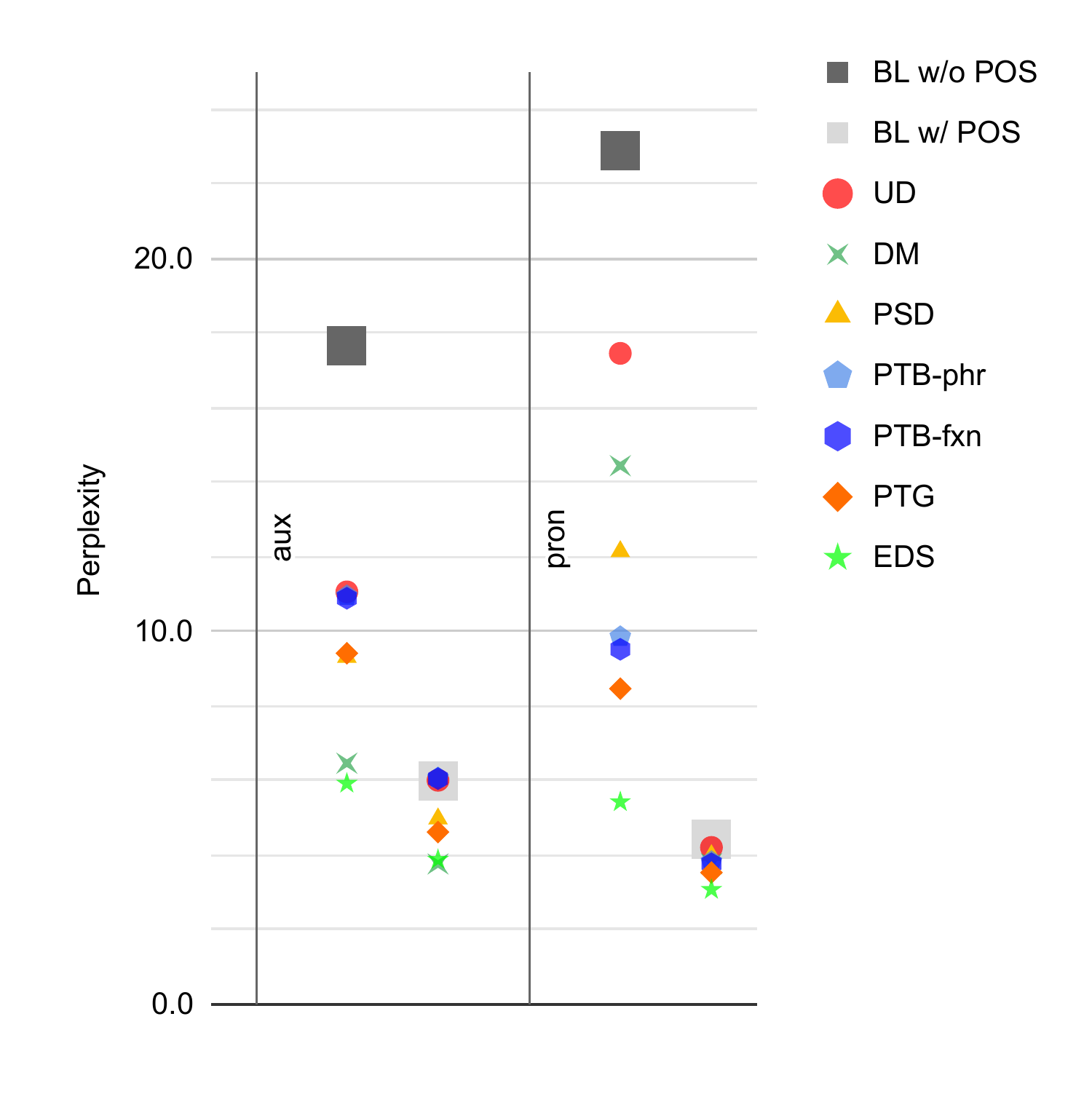}\\
    \includegraphics[width=0.39\textwidth]{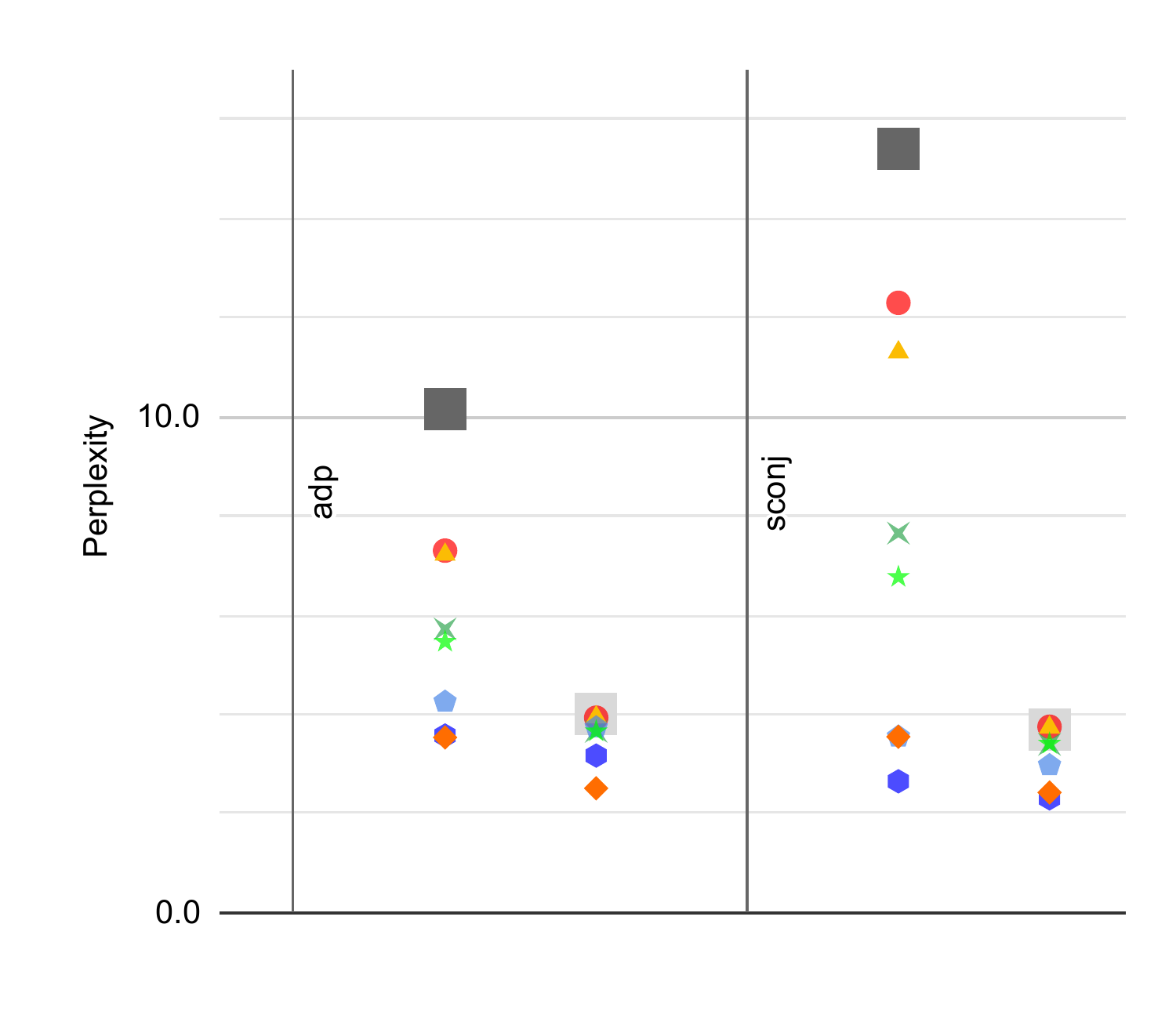}%
    \includegraphics[width=0.39\textwidth]{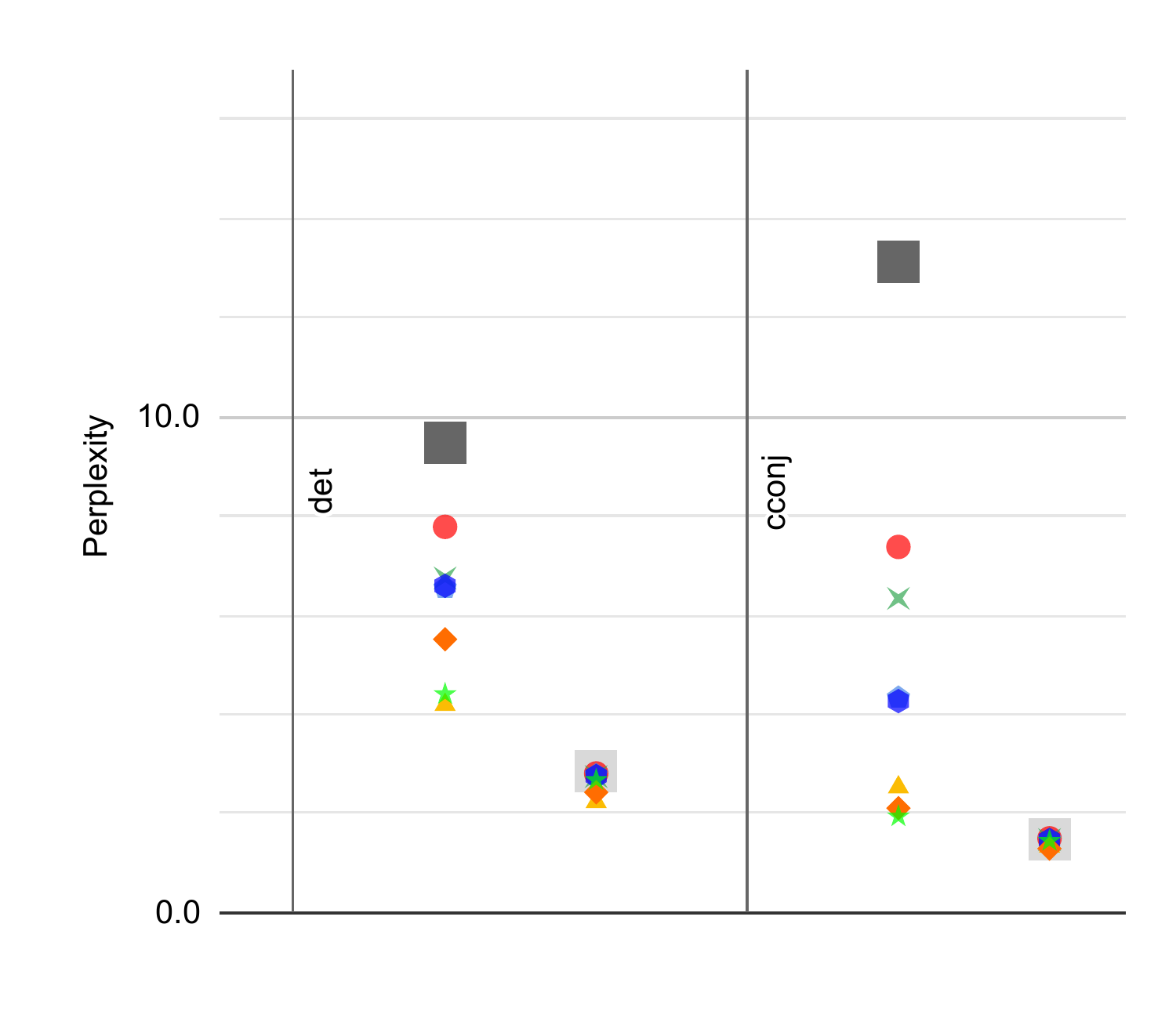}
    \caption{Model perplexity (lower is better) with UPOS as additional input. Top left: nouns, verbs, and modifiers; top right: auxiliaries and pronouns; bottom left: adpositions and subordinating conjunctions; bottom right: determiners and coordinating conjunctions. Big gray squares mark baseline (finetuned GPT-2) performance without (dark) and with (light) POS inputs and SLR-specific data points without/with POS inputs follow below the squares in each respective column. Mind the different y-axis scales, and in particular the log scale in the top-left plot, which makes it easier to read very big and slightly smaller (but still big) differences at the same time.}
    \label{fig:pos-input}
\end{figure*}

A few particularly interesting POS subsets are shown in \cref{fig:pos-input}. We discuss them in order.

Among content words, nouns and verbs are similar both in terms of baseline performance and in how much easier it becomes to select the correct lexical item if the part-of-speech is known. At the same time, the individual SLR formalisms differ quite a lot in how much information they contribute about the POS class itself and about lexical choice within the part-of-speech. The respective best formalisms (EDS for nouns, PTB and UD for verbs) approximate oracle POS knowledge by themselves and still contribute substantial complementary information when the actual POS tag is revealed. In contrast, PTB does not seem to provide any useful signal about nouns to the incremental LM---neither independently nor in conjunction with the POS.

Modifiers (adjectives and adverbs) display a rather interesting behavior: the fact \textit{that} a word of this type is coming next is very hard to predict from just the preceding raw context, which makes sense since they tend to add \textit{optional} meaning on top of the (obligatory) logical and grammatical content. However, once the decision to modify has been made, the contextual choice becomes much easier than that for nouns or verbs. In both cases, all SLRs are quite helpful, with UD on the lower end and EDS leading the field.

We find similar tendencies among auxiliaries ($\approx$function verbs) and pronouns ($\approx$function nouns) as with (content) verbs and (content) nouns, but naturally at a much smaller scale. Despite their functional-grammatical distribution and behavior, the semantic frameworks EDS and PTG consistently outperform the syntactic ones UD and PTB even on these `small' words. A possible explanation for this interaction with auxiliaries in particular could be that EDS and PTG do not analyze them separately at all, but rather group them, respectively, with the preceding context\footnote{EDS, like PSD, actually has no anchors for auxiliaries; we attach them to the preceding semantic unit by default.} or their main predicate. The models might be able to leverage this to focus on things like subject-verb agreement, local cohesion, or anticipating the main predicate. More explicit syntactic analyses of auxiliaries (incrementally inaccessible forward-pointing dependencies in UD; VP-nesting in PTB), in contrast, may restrict the model from directly making these connections.
Adding POS information in the input decreases SLR-dependent differences.

For `subordinators' in the broad sense, i.e., subordinating conjunctions at the clausal level and adpositions for nominal complements, PTB and PTG are particularly well-suited. By themselves they are already \textit{at least} as informative as POS, and they still add a small but noticeable complementary signal when the POS is revealed.

Determiners and coordinating conjunctions, which both already show extremely low perplexity with some SLR models (namely, EDS, PSD, and PTG), entirely lose any reliance on particular SLRs when their POS is known.

\onecolumn

\begin{landscape}

\begin{table*}[p]
    \centering\small
    \begin{tabular}{lrHHr r|rrrrrrr H}
        & \multicolumn{1}{c}{Eval} & & & \multicolumn{1}{c}{Train} &
        \multicolumn{9}{c}{{Perplexity}} \\\cmidrule(lr){2-2}\cmidrule(lr){5-5}\cmidrule(l){6-14}
        \textbf{POS} & \multicolumn{1}{c}{\textbf{Toks}} & \multicolumn{1}{H}{\textbf{Vocab}} & \multicolumn{1}{H}{\textbf{Toks}} & \multicolumn{1}{c}{\textbf{Vocab}} & \multicolumn{1}{c}{\textbf{GPT-2}} & \multicolumn{1}{|c}{\textbf{UD}} & \multicolumn{1}{c}{\textbf{DM}} & \multicolumn{1}{c}{\textbf{PSD}} & \multicolumn{1}{c}{\textbf{PTB-phr}} & \multicolumn{1}{c}{\textbf{PTB-fxn}} & \multicolumn{1}{c}{\textbf{PTG}} & \multicolumn{1}{c}{\textbf{EDS}} & \multicolumn{1}{H}{\textbf{SLR Avg}} \\\toprule
        
All & 22,596 & 5,364 & 658,475 & 27,344 & 45.9	\textcolor{gray}{\scriptsize $\pm \phantom{\ <\ } .1$} & 	32.7	\textcolor{gray}{\scriptsize $\pm \phantom{\ <\ }0.2$} & 	31.4	\textcolor{gray}{\scriptsize $\pm \phantom{\ <\ } 0.1$} & 	30.7	\textcolor{gray}{\scriptsize $\pm \phantom{\ <\ } 0.1$} & 	29.8	\textcolor{gray}{\scriptsize $\pm \phantom{\ <\ } 0.2$} & 	29.0	\textcolor{gray}{\scriptsize $\pm \phantom{\ <\ } 0.3$} & 	26.8	\textcolor{gray}{\scriptsize $\pm \phantom{\ <\ } 0.3$} & 	\textbf{24.7}	\textcolor{gray}{\scriptsize $\pm \phantom{\ <\ } 0.3$} & 	29.3	\textcolor{gray}{\scriptsize $\pm \phantom{0}2.8$}   \\\midrule
noun & 7,731 & 3,130 & 223,267 & 18,435 & 142.5	\textcolor{gray}{\scriptsize $\pm \phantom{\ <\ } .7$} &	122.0	\textcolor{gray}{\scriptsize $\pm \phantom{\ <\ } 0.1$} &	119.0	\textcolor{gray}{\scriptsize $\pm \phantom{\ <\ } 1.0$} &	120.9	\textcolor{gray}{\scriptsize $\pm \phantom{\ <\ } 0.7$} &	138.0	\textcolor{gray}{\scriptsize $\pm \phantom{\ <\ } 2.5$} &	139.6	\textcolor{gray}{\scriptsize $\pm \phantom{\ <\ } 4.5$} &	120.6	\textcolor{gray}{\scriptsize $\pm \phantom{\ <\ } 2.5$} &	\textbf{98.0}	\textcolor{gray}{\scriptsize $\pm \phantom{\ <\ } 3.6$} &	122.6	\textcolor{gray}{\scriptsize $\pm 13.9$} \\
verb & 2,639 & 1,177 & 74,218 & 7,100 & 128.8	\textcolor{gray}{\scriptsize $\pm \phantom{\ <\ } .8$} &	\textbf{80.4}	\textcolor{gray}{\scriptsize $\pm \phantom{\ <\ } 1.2$} &	89.4	\textcolor{gray}{\scriptsize $\pm \phantom{\ <\ } 1.1$} &	90.7	\textcolor{gray}{\scriptsize $\pm \phantom{\ <\ } 0.7$} &	81.7	\textcolor{gray}{\scriptsize $\pm \phantom{\ <\ } 0.9$} &	\textbf{79.3}	\textcolor{gray}{\scriptsize $\pm \phantom{\ <\ } 0.7$} &	86.9	\textcolor{gray}{\scriptsize $\pm \phantom{\ <\ } 2.5$} &	85.9	\textcolor{gray}{\scriptsize $\pm \phantom{\ <\ } 1.4$} &	84.9	\textcolor{gray}{\scriptsize $\pm \phantom{0}4.5$}  \\
mod & 2,235 & 934 & 65,900 & 6,292 & 228.7	\textcolor{gray}{\scriptsize $\pm \phantom{\ <\ } .5$} &	158.8	\textcolor{gray}{\scriptsize $\pm \phantom{\ <\ } 3.8$} &	145.9	\textcolor{gray}{\scriptsize $\pm \phantom{\ <\ } 3.3$} &	130.5	\textcolor{gray}{\scriptsize $\pm \phantom{\ <\ } 2.0$} &	123.4	\textcolor{gray}{\scriptsize $\pm \phantom{\ <\ } 2.1$} &	113.5	\textcolor{gray}{\scriptsize $\pm \phantom{\ <\ } 3.3$} &	\textbf{100.2}	\textcolor{gray}{\scriptsize $\pm \phantom{\ <\ } 4.0$} &	\textbf{98.6}	\textcolor{gray}{\scriptsize $\pm \phantom{\ <\ } 4.0$} &	124.4	\textcolor{gray}{\scriptsize $\pm 22.6$} \\[7pt]

aux & 582 & 42 & 18,303 & 95 & 17.6	\textcolor{gray}{\scriptsize $\pm <.1$} &	11.1	\textcolor{gray}{\scriptsize $\pm \phantom{\ <\ } 0.1$} &	6.5	\textcolor{gray}{\scriptsize $\pm \phantom{\ <\ } 0.1$} &	9.3	\textcolor{gray}{\scriptsize $\pm \phantom{\ <\ } 0.2$} &	11.0	\textcolor{gray}{\scriptsize $\pm \phantom{\ <\ } 0.2$} &	10.9	\textcolor{gray}{\scriptsize $\pm \phantom{\ <\ } 0.1$} &	9.4	\textcolor{gray}{\scriptsize $\pm \phantom{\ <\ } 0.1$} &	\textbf{5.9}	\textcolor{gray}{\scriptsize $\pm \phantom{\ <\ } 0.1$} &	9.1	\textcolor{gray}{\scriptsize $\pm \phantom{0} 2.1$} \\
adp & 1,957 & 88 & 59,618 & 232 & 10.1	\textcolor{gray}{\scriptsize $\pm <.1$} &	7.3	\textcolor{gray}{\scriptsize $\pm \phantom{\ <\ } 0.1$} &	5.7	\textcolor{gray}{\scriptsize $\pm \phantom{\ <\ } 0.1$} &	7.2	\textcolor{gray}{\scriptsize $\pm \phantom{\ <\ } 0.1$} &	4.3	\textcolor{gray}{\scriptsize $\pm <0.1$} &	\textbf{3.6}	\textcolor{gray}{\scriptsize $\pm <0.1$} &	\textbf{3.5}	\textcolor{gray}{\scriptsize $\pm \phantom{\ <\ } 0.1$} &	5.5	\textcolor{gray}{\scriptsize $\pm \phantom{\ <\ } 0.1$} &	5.3	\textcolor{gray}{\scriptsize $\pm \phantom{0} 1.6$} \\
part & 645 & 17 & 17,263 & 27 & 3.7	\textcolor{gray}{\scriptsize $\pm <.1$} &	2.0	\textcolor{gray}{\scriptsize $\pm <0.1$} &	2.1	\textcolor{gray}{\scriptsize $\pm \phantom{\ <\ } 0.1$} &	2.3	\textcolor{gray}{\scriptsize $\pm \phantom{\ <\ } 0.1$} &	\textbf{1.7}	\textcolor{gray}{\scriptsize $\pm \phantom{\ <\ } 0.1$} &	1.7	\textcolor{gray}{\scriptsize $\pm <0.1$} &	1.7	\textcolor{gray}{\scriptsize $\pm <0.1$} &	\textbf{1.6}	\textcolor{gray}{\scriptsize $\pm <0.1$} &	1.9	\textcolor{gray}{\scriptsize $\pm \phantom{0} 0.3$} \\
sconj & 268 & 42 & 9,053 & 96 & 15.4	\textcolor{gray}{\scriptsize $\pm \phantom{\ <\ } .1$} &	12.3	\textcolor{gray}{\scriptsize $\pm \phantom{\ <\ } 0.1$} &	7.7	\textcolor{gray}{\scriptsize $\pm \phantom{\ <\ } 0.2$} &	11.3	\textcolor{gray}{\scriptsize $\pm \phantom{\ <\ } 0.6$} &	3.5	\textcolor{gray}{\scriptsize $\pm <0.1$} &	\textbf{2.6}	\textcolor{gray}{\scriptsize $\pm <0.1$} &	3.5	\textcolor{gray}{\scriptsize $\pm \phantom{\ <\ } 0.1$} &	6.8	\textcolor{gray}{\scriptsize $\pm \phantom{\ <\ } 0.1$} &	6.8	\textcolor{gray}{\scriptsize $\pm \phantom{0} 3.9$} \\
cconj & 548 & 10 & 14,485 & 35 & 13.0	\textcolor{gray}{\scriptsize $\pm \phantom{\ <\ } .1$} &	7.4	\textcolor{gray}{\scriptsize $\pm \phantom{\ <\ } 0.3$} &	6.3	\textcolor{gray}{\scriptsize $\pm \phantom{\ <\ } 0.5$} &	2.5	\textcolor{gray}{\scriptsize $\pm \phantom{\ <\ } 0.1$} &	4.3	\textcolor{gray}{\scriptsize $\pm \phantom{\ <\ } 0.1$} &	4.3	\textcolor{gray}{\scriptsize $\pm \phantom{\ <\ } 0.1$} &	2.1	\textcolor{gray}{\scriptsize $\pm <0.1$} &	\textbf{1.9}	\textcolor{gray}{\scriptsize $\pm <0.1$} &	4.1	\textcolor{gray}{\scriptsize $\pm \phantom{0} 2.1$} \\
det & 1,726 & 35 & 51,762 & 91 & 9.4	\textcolor{gray}{\scriptsize $\pm \phantom{\ <\ } .1$} &	7.8	\textcolor{gray}{\scriptsize $\pm \phantom{\ <\ } 0.1$} &	6.7	\textcolor{gray}{\scriptsize $\pm \phantom{\ <\ } 0.4$} &	\textbf{4.2}	\textcolor{gray}{\scriptsize $\pm \phantom{\ <\ } 0.1$} &	6.5	\textcolor{gray}{\scriptsize $\pm \phantom{\ <\ } 0.4$} &	6.6	\textcolor{gray}{\scriptsize $\pm \phantom{\ <\ } 0.4$} &	5.5	\textcolor{gray}{\scriptsize $\pm \phantom{\ <\ } 0.6$} &	\textbf{4.4}	\textcolor{gray}{\scriptsize $\pm \phantom{\ <\ } 0.2$} &	6.0	\textcolor{gray}{\scriptsize $\pm \phantom{0} 1.3$} \\
pron & 868 & 68 & 23,473 & 149 & 22.9	\textcolor{gray}{\scriptsize $\pm <.1$} &	17.5	\textcolor{gray}{\scriptsize $\pm \phantom{\ <\ } 0.2$} &	14.4	\textcolor{gray}{\scriptsize $\pm \phantom{\ <\ } 0.3$} &	12.1	\textcolor{gray}{\scriptsize $\pm <0.1$} &	9.9	\textcolor{gray}{\scriptsize $\pm \phantom{\ <\ } 0.5$} &	9.5	\textcolor{gray}{\scriptsize $\pm \phantom{\ <\ } 0.1$} &	8.5	\textcolor{gray}{\scriptsize $\pm \phantom{\ <\ } 0.1$} &	\textbf{5.4}	\textcolor{gray}{\scriptsize $\pm \phantom{\ <\ } 0.1$} &	11.0	\textcolor{gray}{\scriptsize $\pm \phantom{0} 4.0$} \\[7pt]

num & 719 & 200 & 26,161 & 1,059 & 72.6	\textcolor{gray}{\scriptsize $\pm \phantom{\ <\ } .3$} &	57.1	\textcolor{gray}{\scriptsize $\pm \phantom{\ <\ } 0.7$} &	55.2	\textcolor{gray}{\scriptsize $\pm \phantom{\ <\ } 0.6$} &	53.3	\textcolor{gray}{\scriptsize $\pm \phantom{\ <\ } 0.5$} &	58.4	\textcolor{gray}{\scriptsize $\pm \phantom{\ <\ } 0.6$} &	58.6	\textcolor{gray}{\scriptsize $\pm \phantom{\ <\ } 2.5$} &	\textbf{48.3}	\textcolor{gray}{\scriptsize $\pm \phantom{\ <\ } 0.4$} &	\textbf{47.5}	\textcolor{gray}{\scriptsize $\pm \phantom{\ <\ } 1.0$} &	54.1	\textcolor{gray}{\scriptsize $\pm \phantom{0} 4.6$} \\
punct & 2,527 & 31 & 69,358 & 68 & 4.9	\textcolor{gray}{\scriptsize $\pm <.1$} &	\textbf{2.3}	\textcolor{gray}{\scriptsize $\pm <0.1$} &	2.8	\textcolor{gray}{\scriptsize $\pm <0.1$} &	3.2	\textcolor{gray}{\scriptsize $\pm <0.1$} &	\textbf{2.3}	\textcolor{gray}{\scriptsize $\pm <0.1$} &	\textbf{2.4}	\textcolor{gray}{\scriptsize $\pm \phantom{\ <\ } 0.1$} &	2.6	\textcolor{gray}{\scriptsize $\pm \phantom{\ <\ } 0.1$} &	2.7	\textcolor{gray}{\scriptsize $\pm <0.1$} &	2.6	\textcolor{gray}{\scriptsize $\pm \phantom{0} 0.3$} \\
misc & 151 & 24 & 5,614 & 183 & 7.0	\textcolor{gray}{\scriptsize $\pm <.1$} &	4.6	\textcolor{gray}{\scriptsize $\pm \phantom{\ <\ } 0.1$} &	5.1	\textcolor{gray}{\scriptsize $\pm \phantom{\ <\ } 0.5$} &	5.2	\textcolor{gray}{\scriptsize $\pm \phantom{\ <\ } 0.3$} &	\textbf{3.7}	\textcolor{gray}{\scriptsize $\pm \phantom{\ <\ } 0.1$} &	\textbf{3.5}	\textcolor{gray}{\scriptsize $\pm \phantom{\ <\ } 0.2$} &	5.6	\textcolor{gray}{\scriptsize $\pm \phantom{\ <\ } 0.2$} &	\textbf{4.0}	\textcolor{gray}{\scriptsize $\pm \phantom{\ <\ } 0.4$} &	4.5	\textcolor{gray}{\scriptsize $\pm \phantom{0} 0.8$} \\
\bottomrule
    \end{tabular}
    \caption{PPL breakdown by UPOS classes of individual models (3-seed-averages), and the macro-average $\pm$ stdev over all SLR-combined models. We show token counts and observed vocabulary sizes for reference. mod---adjectives and adverbs; aux---auxiliary verbs; adp---adpositions; part---particles; sconj---subordinating conjunctions; cconj---coordinating conjunctions; det---determiners; pron---pronouns; num---numbers; punct---punctuation. Best results (within the variance) in each row are \textbf{bolded}.}
    \label{tab:pos_results_ext}
\end{table*}

\begin{table*}[p]
    \centering\small
    \begin{tabular}{lc rrrrrrr}
         \multicolumn{1}{c}{\textbf{Ablation}} & \multicolumn{1}{c}{\textbf{Applied in}} & \multicolumn{1}{c}{\textbf{UD}} & \multicolumn{1}{c}{\textbf{DM}} & \multicolumn{1}{c}{\textbf{PSD}} & \multicolumn{1}{c}{\textbf{PTB-phr}} & \multicolumn{1}{c}{\textbf{PTB-fxn}} & \multicolumn{1}{c}{\textbf{PTG}} & \multicolumn{1}{c}{\textbf{EDS}} \\\toprule
\textit{Full} &  & 32.7  & 31.4  & 30.7  & 29.8  & 29.0  & 26.8  & 24.7   \\\midrule
$\neovsearrow$ Labels & testing & $+$10.2 & $+$4.7 & $+$5.8 & $+$63.0 & $+$73.9 & $+$18.7 & $+$21.8 \\
$\neovsearrow$ Anchors & testing & $+$110.8 & $+$34.8 & $+$22.4 & $+$177.0 & $+$223.1 & $+$103.4 & $+$70.4 \\
$\neovsearrow$ Both & testing & $+$92.4 & $+$33.4 & $+$19.2 & $+$168.9 & $+$207.4 & $+$80.1 & $+$70.1 \\[7pt]
$\neovsearrow$ Labels & training & $+$2.0 & $+$1.4 & $+$1.6 & $+$8.4 & $+$9.0 & $+$5.3 & $+$1.9 \\
$\neovsearrow$ Anchors & training & $+$13.0 & $+$8.5 & $+$6.3 & $+$16.9 & $+$17.5 & $+$18.1 & $+$12.8 \\
$\neovsearrow$ Both & training & $+$13.0 & $+$7.8 & $+$6.1 & $+$17.3 & $+$18.3 & $+$18.5 & $+$12.8 \\[7pt]
$\neovsearrow$ Labels & both & $+$2.0 & $+$1.3 & $+$1.7 & $+$8.4 & $+$9.3 & $+$5.4 & $+$2.0 \\
$\neovsearrow$ Anchors & both & $+$13.1 & $+$7.9 & $+$6.0 & $+$16.7 & $+$17.5 & $+$19.2 & $+$14.0 \\
$\neovsearrow$ Both & both & $+$13.3 & $+$7.3 & $+$6.0 & $+$16.8 & $+$18.1 & $+$19.4 & $+$14.1 \\
        $-$ SLR & both & $+$13.2  & $+$14.5  & $+$15.2  & $+$16.1  & $+$16.9  & $+$19.1  & $+$21.2   \\
        \bottomrule
    \end{tabular}
    \caption{Ablations measured in absolute perplexity difference ($\Delta$PPL). 
    \textit{Full} and $-$SLR correspond, respectively, to \cref{tab:main_results}'s rows 4 (DM) \slash\ 6 (PTB-phr) and row 2 (GPT-2 $+$Domain).}
    \label{tab:abl_results_ext}
\end{table*}

\end{landscape}

\end{document}